\documentclass[journal]{IEEEtran}
\usepackage{algorithm}
\usepackage{algorithmic}
\usepackage{graphicx}
\usepackage{epstopdf}
\usepackage{cite}
\usepackage{makecell}
\usepackage{array}
\usepackage{multirow}
\usepackage{amsfonts,amssymb}
\usepackage{amsmath}
\usepackage{flushend}

\usepackage{algorithm}
\usepackage{algorithmic}
\usepackage{booktabs}

\begin{document}
\newcolumntype{R}[1]{>{\raggedleft\arraybackslash}p{#1}}
\newcolumntype{L}[1]{>{\raggedright\arraybackslash}p{#1}}
\newcolumntype{C}[1]{>{\centering\arraybackslash}p{#1}}

%
\title{Automatic Design of CNNs via Differentiable Neural Architecture Search for PolSAR Image Classification}
\author{Hongwei~Dong,
        Siyu~Zhang,
        Bin~Zou,~\IEEEmembership{Senior Member,~IEEE},
        Lamei~Zhang,~\IEEEmembership{Senior Member,~IEEE,}
\thanks{This work was supported in part by the National Natural Science Foundation of China (61401124, 61871158), in part by Scientific Research Foundation for the Returned Overseas Scholars of Heilongjiang Province (LC2018029), in part by Aeronautical Science Foundation of China (20182077008). (\emph{Corresponding author: Lamei Zhang.})}
\thanks{The authors are with the Department of Information Engineering, Harbin Institute of Technology, Harbin 150001, China (e-mail: donghongwei1994@163.com; lmzhang@hit.edu.cn;).}}

\markboth{}
{Shell \MakeLowercase{\textit{et al.}}: Bare Demo of IEEEtran.cls for IEEE Journals}

\maketitle

\begin{abstract}
Convolutional neural networks (CNNs) have shown good performance in polarimetric synthetic aperture radar (PolSAR) image classification due to the automation of feature engineering. Excellent hand-crafted architectures of CNNs incorporated the wisdom of human experts, which is an important reason for CNN's success. However, the design of the architectures is a difficult problem, which needs a lot of professional knowledge as well as computational resources. Moreover, the architecture designed by hand might be suboptimal, because it is only one of thousands of unobserved but objective existed paths. Considering that the success of deep learning is largely due to its automation of the feature engineering process, how to design automatic architecture searching methods to replace the hand-crafted ones is an interesting topic. In this paper, we explore the application of neural architecture search (NAS) in PolSAR area for the first time. Different from the utilization of existing NAS methods, we propose a differentiable architecture search (DAS) method which is customized for PolSAR classification. The proposed DAS is equipped with a PolSAR tailored search space and an improved one-shot search strategy. By DAS, the weights parameters and architecture parameters (corresponds to the hyperparameters but not the topologies) can be optimized by stochastic gradient descent method during the training. The optimized architecture parameters should be transformed into corresponding CNN architecture and re-train to achieve high-precision PolSAR classification. In addition, complex-valued DAS is developed to take into account the characteristics of PolSAR images so as to further improve the performance. Experiments on three PolSAR benchmark datasets show that the CNNs obtained by searching have better classification performance than the hand-crafted ones.
\end{abstract}

\begin{IEEEkeywords}
Polarimetric synthetic aperture radar (PolSAR) classification, convolutional neural network (CNN), neural architecture search (NAS), Automatic machine learning (AutoML)
\end{IEEEkeywords}

\IEEEpeerreviewmaketitle

\section{Introduction}
\IEEEPARstart{P} {olarimetric} synthetic aperture radar (PolSAR) is an indispensable sensor in earth observation. It can provide abundant target information in all-weather and all-times, which is not available for other sensors. More and more attention has been paid to the development PolSAR and the interpretation of PolSAR images. As the basis of PolSAR interpretation, PolSAR image classification has been a hot research field in recent years. For quite a long time in the past, features with clear physical scattering meanings are eager to be studied in PolSAR classification area, i.e. target decomposition techniques. The idea of target decomposition is to decompose PolSAR image into several known scattering characteristics. The earliest Pauli decomposition is based on polarization scattering matrix by defining different basis matrix. Then many target decomposition based feature extraction methods have been introduced, such as Cameron decomposition \cite{oldmethod2}, Yamaguchi four-component scattering power decomposition \cite{oldmethod3}, and some variants \cite{oldmethod1}. Statistical based feature methods have also been widely used and some unsupervised feature engineering methods \cite{halpha,PCA} are explored to realize better classification results. However, it is still difficult to construct complete features due to the complex imaging mechanism of PolSAR. Therefore, PolSAR image classification is still an open problem.
\par Deep learning \cite{Deep} has made remarkable progress in natural language processing, computer vision and it has the potential to be applied in more fields. Its main idea is to automatically learn features through a neural network instead of hand-crafted feature filters. As the main carrier of deep learning technique in image processing field, convolutional neural network (CNN) has aroused the interest of a large number of researchers since AlexNet and ImageNet Large-Scale Visual Recognition Challenge 2012 \cite{Krizhevsky2012ImageNet}. Compared with the ordinary CNN \cite{Lecun2014Backpropagation}, AlexNet explored its ability to process large-scale data by increasing the depth of the network. Given AlexNet's great success, a large number of studies have been carried out to design powerful architectures so as to obtain better performance \cite{vgg,Szegedy2014Going,Highway,ResNet,DenseNet,xception,resnext,senet}. Novel architectures have greatly promoted the development of CNNs. It can be said that CNN's current achievements are largely due to the efforts of experts on the issue of neural architecture design.
\par Some researchers in SAR and PolSAR area have begun to use CNNs to improve image classification due to the difficulty in designing features manually. Geng \emph{et al.} firstly explored the application of deep neural network in SAR image classification \cite{7286736}. After that, many CNN-based classification methods have sprung up. They can be divided into two parts according to different research objectives: target recognition and terrain classification. Standard SAR image automatic target recognition (SAR-ATR) can be divided into three stages: detection, discrimination and classification. The last step is very important, which needs a fine recognition of the image slices with possible targets. In order to achieve high-precision target recognition, an all-convolutional architecture was proposed in \cite{7460942} for SAR-ATR. Over $99\%$ accuracy was obtained on Moving and Stationary Target Acquisition and Recognition (MSTAR) dataset. A Highway units based architecture was proposed in \cite{7926358} to achieve SAR image target recognition in the environment of limited training samples. They obtained the matched precision compared with the method in \cite{7460942} under the premise of reducing $30\%$ training data. Multiview SAR data was generated and a multi-input CNN architecture was proposed in \cite{8207785} to get the features of targets from different views. In order to reduce the demand of extensive labeled samples, high-order features extracted by CNN were made into feature dictionary, and the end-to-end training was carried out through feature metric and two-stage optimization \cite{8375097}. Different from the task of target recognition, the purpose of terrain classification is to predict each pixel of a PolSAR image. At this stage, the general paradigm of PolSAR image terrain classification follows the first work \cite{7762055}: The complex-valued PolSAR data are first converted into a real-valued tensor. Well-designed CNNs are then used for classification. The architecture with two convolutional layers was proposed in \cite{7762055} and achieved $92.46\%$ accuracy on PolSAR Flevoland data. Transfer learning based ensemble CNNs was used to realize scene classification on the data acquired from P-band Chinese airborne PolSAR system \cite{8363061}. In order to make up the shortages of slice-based classification methods, i.e. slow speed and lack of global information, fully convolutional networks \cite{Shelhamer2014Fully} was introduced to achieve better full image classification \cite{Yan2018A,8601351}. Some works studied new representations of PolSAR data, so as to improve the performance of CNN-based methods from the perspective of data \cite{input1,input2,input3}. Considering the particularity of PolSAR data, complex-valued CNN architectures have also been studied to some extent \cite{complex,complex2,complex3}. At present, the research of CNN-based PolSAR image classification methods mainly focus on building PolSAR tailored network architectures, or designing architectures for the environment of limited training samples.
\par As can be seen from the work described above: Manually designing CNN's architecture is the problem that researchers have been engaged in the past few years, whether in optical image or radar image area. After a lot of research, there are many experiences and methods to guide the design of neural architectures. Therefore, a new research direction is emerging, i.e. neural architecture search (NAS), to replace the classical hand-crafted architectures. The superiority of NAS is reflected in two aspects. On the one hand, the success of deep learning is largely due to its automation of the feature engineering process. It can be expected that if this automated process is extended to the machine learning model design stage, it will be possible to further improve its performance. On the flip side, it is a costly task to design a tailored neural architecture for a certain problem. This is also the biggest difficulty in utilizing deep learning technique to solve problems in other fields, e.g. radar image processing. Zoph \emph{et al.} proposed the concept of NAS to deal with these two problems \cite{zophrlnas}. They noticed that the topology and hyperparameter of a neural network could be typically specified by a variable-length string. Therefore, a recurrent network called controller was used to generate such string. The controller was trained with reinforcement learning \cite{rl} to maximize the expected accuracy of the string-transformed architecture on validation set. Theoretically, one can get the most powerful architecture in the current search space at the end of the controller's training. This work has some overlap with automatic machine learning \cite{automl} and meta learning \cite{meta,metalearning}, and it also lays a foundation for reinforcement learning based NAS studies. Then, improvements for such kind of search methods began to emerge. Zoph \emph{et al.} searched architectures on a small dataset and then transferred it to large dataset, called NASNet, achieved the state-of-the-art on ImageNet dataset \cite{8579005}. NASNet got $1.2\%$ better in top-1 accuracy than the best hand-crafted architecture while having lower computational complexity. The powerful ability of NAS attracts a large number of researchers, and many different search strategies were used to search better architectures, such as random search \cite{random}, Bayesian optimization and evolutionary methods \cite{evnas,Liua}. However, the methods based on evolution or reinforcement learning require training all candidates over and over again to find the optimal architecture, which brings unaffordable computational difficulties. For example, one needed to spend $10^4$ GPU days in order to obtain a good neural architecture \cite{8579005,Real2018}, which seriously restricts the development of NAS's theory and application. Therefore, many acceleration methods have been studied, including small but effective cells-based search space \cite{Liua}, fast model evaluation \cite{Elsken2019,cai2019} and proxy mechanism \cite{8579005}. Although these methods sped up the search process to some extent, the main reason for their inefficiency was not changed: NAS is treated as discrete sequential decision process, which leads to architecture sampling and evaluation \cite{Rudder}. In order to fundamentally find a low-cost way, Liu \emph{et al.} relaxed the architecture representation and proposed the gradient-based NAS (DARTS) \cite{darts}. Unlike ordinary methods of applying evolution or reinforcement learning with a controller over a discrete and non-differentiable search space, DARTS built a continuous search space by a serious of attention-based architecture parameters. Thanks to DARTS, the search time reduced from thousands of GPU days to one day. However, the approximate solution to the bilevel optimization problem made DARTS unstable. On basis of DARTS, some improved gradient-based methods are being gradually explored \cite{dnas,snas}. It is worth to note that NAS is closely related to network pruning \cite{grouplasso,darc}. The author thinks that the difference between the two lies in: NAS has a considerable search space and its main objective is to find the architecture with high precision in the search space. With a slight difference, pruning methods only search in a small space. They try to find an lightweight architecture while maintaining the accuracy as much as possible. Even so, some ideas shown in pruning methods are of great significance to the development of NAS.
\par The objective of this paper is to improve PolSAR image terrain classification with the aid of CNNs. Based on the previous analysis, the obstacle to achieve this goal lies in the difficulty in designing PolSAR tailored CNN architectures. This difficulty is partly reflected in the fact that architecture design requires considerable expertise of neural networks, which is lacking in PolSAR area. However, the author believes that with the efforts of researchers \cite{complex,input3}, this will not be a stumbling block for future development. The author thinks that the main problem in the future research of CNNs-based PolSAR image classification methods is that the unsatisfactory performance of hand-crafted suboptimal architectures. Therefore, we consider seeking help from NAS to achieve the objective. Based on the above considerations, our objective is transformed to design a cost-effective NAS method to automatically search high-performance architectures, and use the searched architectures to better classify the terrains of PolSAR images.
\par Inspired by previous works, in this paper, a differentiable gradient-based architecture search method (DAS) is proposed according to the characteristics of PolSAR classification to seek the optimal CNN architecture for better terrain classification. More specifically, a PolSAR tailored search space is defined whose variables are the hyperparameters of convolutions, i.e. spatial size and output depth. The reason for such search space comes from the observation that the influence of topology on performance is less than that of hyperparameter setting in the task of image slices based classification. Different from the evolution or reinforcement learning based methods, continuous relaxation is applied to the PolSAR tailored search space, and gradient-based search strategy is implemented. Therefore the search process of DAS is one-shot based on stochastic gradient descent, which brings about fast search speed. Moreover, an improved search strategy is proposed to avoid the repeated convolution operations as well as the approximate bilevel optimization in the classical gradient-based NAS method \cite{darts}. To the best of our knowledge, this is the first work which tries to look for the CNN architecture through automatic search instead of to design it manually, in radar image processing area. Compared with the hand-crafted architectures, the proposed method has better interpretability and stronger performance. Furthermore, complex-valued version of DAS is proposed to fit the complex-valued data format of PolSAR images. It is noteworthy that the proposed methodd can not only be used for searching suitable architectures for PolSAR image classification from scratch, but also improving the existing methods in a plug-and-play manner. Experimental results on three widely used PolSAR benchmark datasets demonstrate the validity of the proposed methods.
\par The rest of this paper is organized as follows: The proposed architecture search method and its complex-valued version are introduced in Section \ref{sec:3}. Experiments and analyses are presented in Section \ref{sec:4}. Section \ref{sec:5} concludes of this paper and gives possible future directions.

\section{Methodology}
\label{sec:3}
In this section, an automatic search method for optimized CNN architecture and its complex-valued version are proposed. The proposed methods are equipped with a PolSAR tailored search space and an improved search strategy, so as to realize a cost-effective search of suitable architectures for PolSAR classification.
\subsection{CNNs for PolSAR Classification}
\begin{figure}[!h]
\begin{centering}
\includegraphics[width=7.5cm]{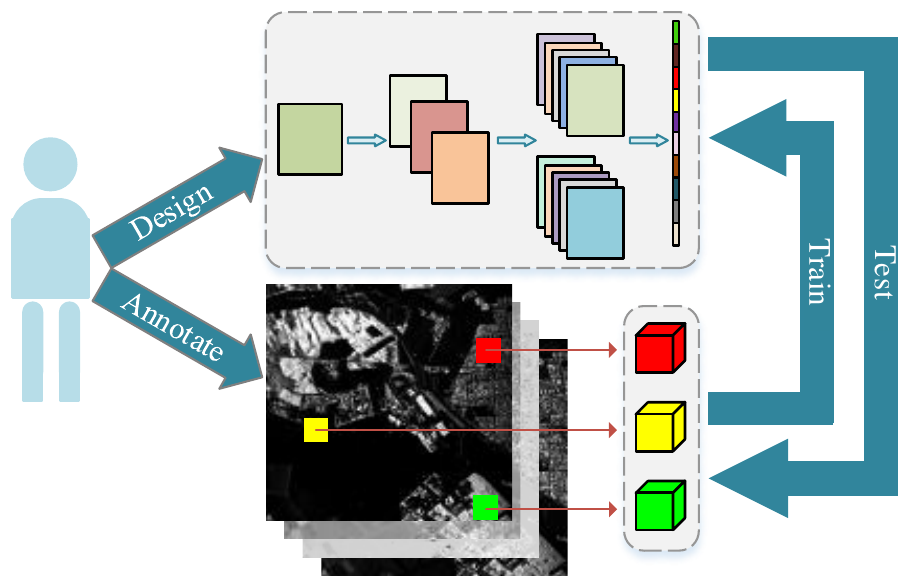}
\caption{General flow chart of classical CNNs-based PolSAR image classification methods.}\label{fig:1}
\end{centering}
\end{figure}

Powerful data fitting ability of CNNs has been valued with the continuous increase of available data. At the beginning, CNN was used for handwritten data recognition \cite{Lecun2014Backpropagation}, and then it became brilliant in image processing \cite{Krizhevsky2012ImageNet,ResNet}. Considering that CNNs have achieved the state-of-the-arts in many tasks of optical image processing, it was applied to radar image processing for the first time to realize high-precision terrain classification \cite{7286736,7762055}. A general paradigm of CNNs-based PolSAR classification methods can be seen from Fig. \ref{fig:1}.
\par As shown in the above figure, the aim of PolSAR image classification is to give a certain category to every pixel. Firstly, small slices are cut around the central pixel according to the polarization coherence matrix of PolSAR data and the ground truth map. Each image slice is treated as a sample with supervisory information and the training and testing sets can be obtained from the labeled image slices. Finally, researchers need to design a suitable CNN architecture and implement training and testing on different sets.

\subsection{Principle of NAS}
CNN uses cascaded feature extractors which are learned (i.e. optimized) from training data. The automation of feature engineering is recognized as an important reason for CNN's success. Accompanied by this success is the transformation from hand-crafted features to hand-crafted neural architectures. This demand led to the birth of NAS \cite{zophrlnas}, which aims to automate the process of architecture engineering.
\par Designing suitable architectures can be treated as a search problem, like hyperparameter optimization \cite{ho}. Search space, search strategy and performance estimation strategy are the components of NAS \cite{surveynas}. Search space is a collection of all possible neural architectures. Search strategy can be seen as the core part of NAS, which is used to find suitable architectures in search space. Every time a candidate architecture is found, it needs to be evaluated by an performance estimation strategy, and loss and accuracy are commonly used evaluation criteria. Decision of the above three comes down to the trade-off between precision and search speed.
\subsection{The Proposed DAS}
We propose a novel architecture search method, named DAS, according to the characteristics of slice-based PolSAR terrain classification. The aim of DAS is to search the hyperparameters, including the size of kernels and the depth of output, of a particular topology. In this work, topology is fixed to a cascaded five-layers convolution network with two pooling layers. The reason for this will be analyzed later. Another aim of DAS is to search the architecture efficiently, so it draws lessons from one-shot NAS methods to a great extent. In DAS, the architectures with different hyperparameter combinations are represented by a series of attention-based architecture parameters. Through stochastic gradient descent on the training set, the architecture parameters can be efficiently optimized, and then the optimized architecture can be obtained by the optimal architecture parameters. The proposal of DAS can be divided into the following steps: Definition of PolSAR tailored search space, continuous relaxation of the search space, linear combination of convolutions, and improved one-shot architecture search.

\subsubsection{PolSAR Tailored Search Space}
\par
\begin{figure}[!h]
\begin{minipage}{0.48\linewidth}
  \centerline{\includegraphics[width=5.0cm]{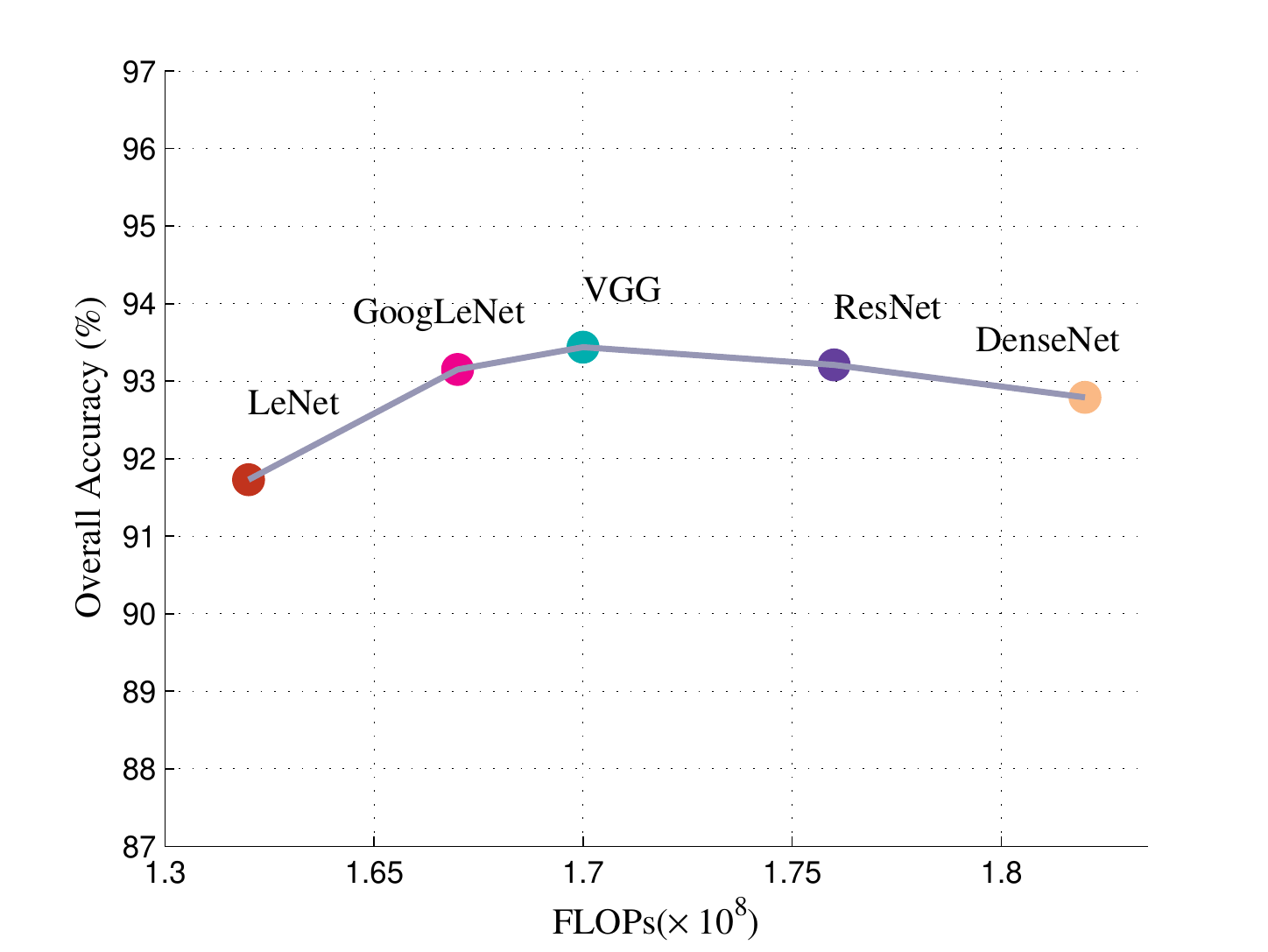}}
  \centerline{(a)}
\end{minipage}
\hfill
\begin{minipage}{0.48\linewidth}
  \centerline{\includegraphics[width=5.0cm]{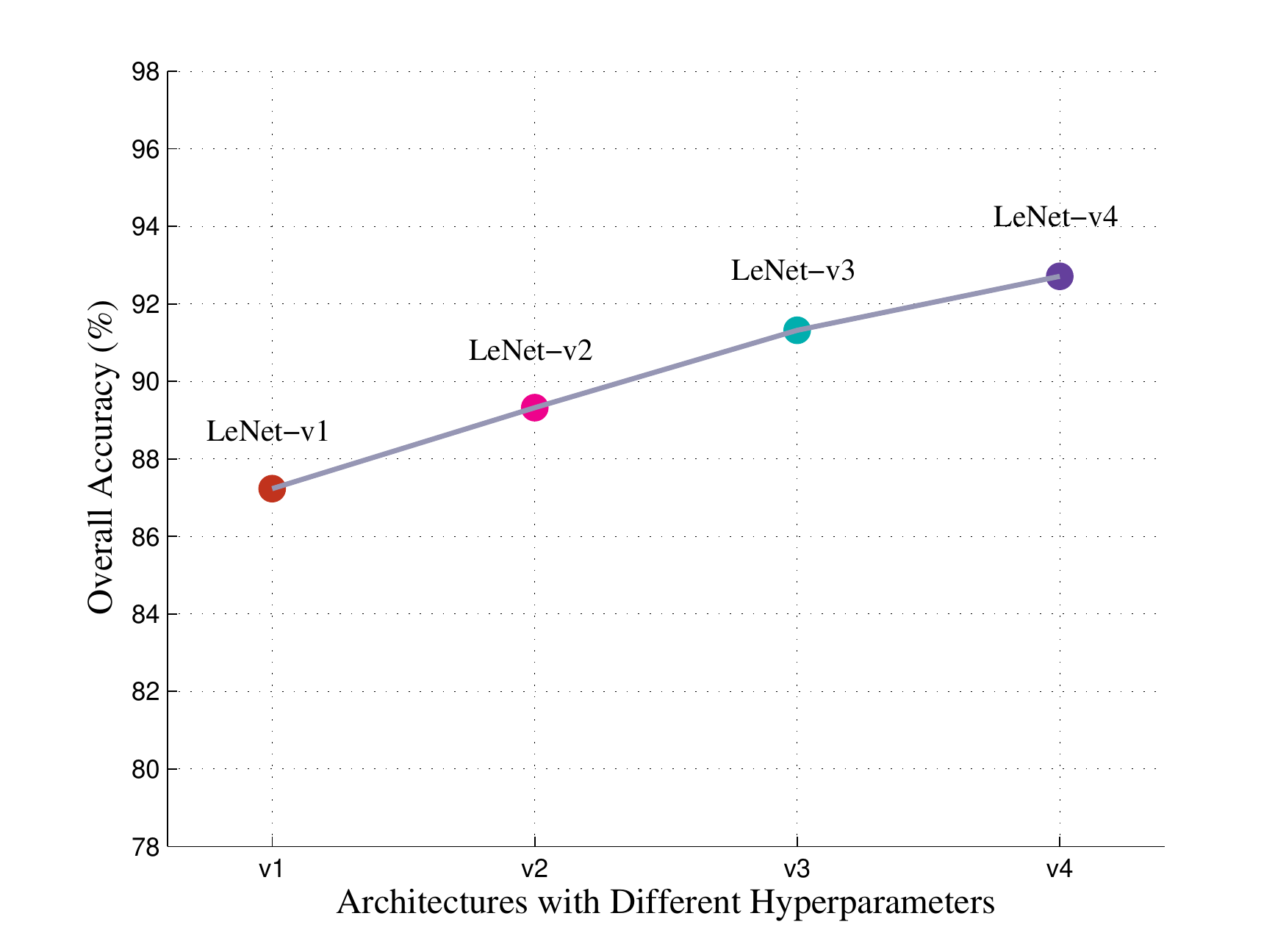}}
  \centerline{(b)}
\end{minipage}
\caption{Study on the change of performance caused by topology and hyperparameter. (a) Performance of the architectures with different topologies but similar computational complexity. (b) Performance of the architectures with the same topology but multiplied number of trainable parameters.}
\label{fig:3}
\end{figure}

 Definition of the search space is important for a search method, because search space determines the number of architectures that can be represented in principle. In general, the search space is exponentially large or even unbounded for a NAS method, which greatly limits the search speed. In fact, the redundancy in search space can be eliminated by incorporating prior knowledge. Considering the fact that slice-based PolSAR image classification is quite different from optical image classification, i.e. inputs of the former is image slices with very small size (usually no more than $20\times 20$), and input size of the latter is generally more than $200\times 200$. For this reason, architectures of CNNs used for PolSAR classification have a shallower depth (less than 10 layers and no more than 3 pooling) compared with optical classification used ones (more than 100 layers). More layers means that the complexity of the topology can grow exponentially, so the search of topologies is popular in the NAS for optical image processing. However, the size of the input image slices limits the depth of PolSAR classification used architectures. Therefore, the following assumption can be made: For image slices based PolSAR classification, the choice of hyperparameters is at least as important as the topologies to the performance of the architectures with similar number of trainable parameters. In order to verify the assumption, numerical simulation is carried out. One side, architectures with different topologies but similar number of trainable weights were trained and tested. The other side, architectures obtained by four kinds of different hyperparameters under the LeNet topology were tested. The experimental results are shown in Fig. \ref{fig:3}, which can support our assumption.
\par
\begin{figure*}[t]
\begin{centering}
\includegraphics[width=16.5cm]{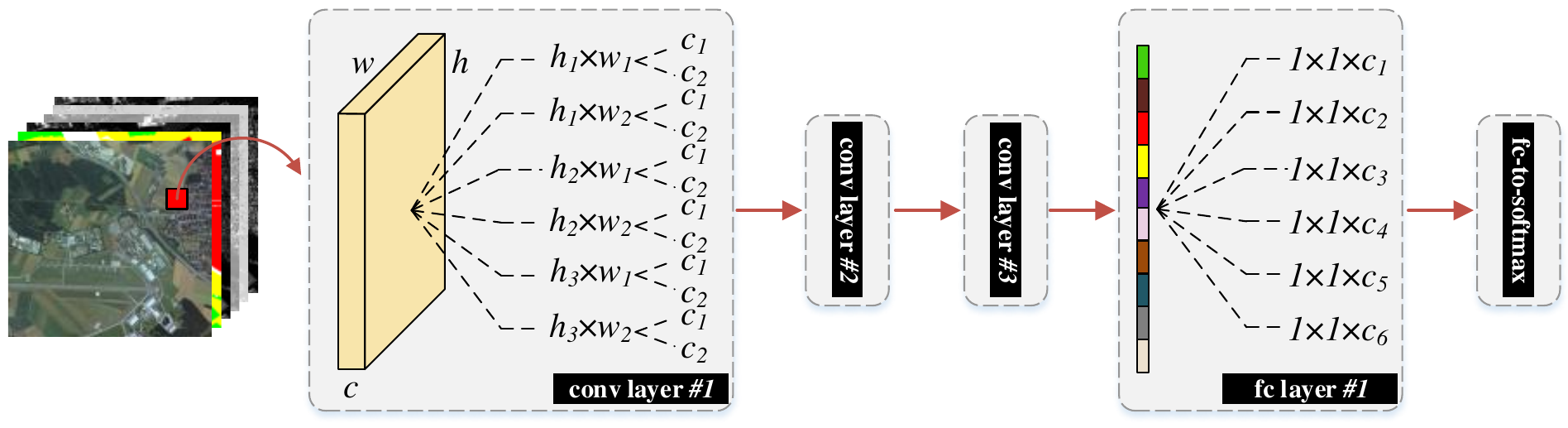}
\caption{Illustration of the proposed PolSAR tailored search space for DAS. Architecture search is carried out based on a five-layers CNN. The variables in the proposed search space are hyperparameters of each trainable layer, not the topology. }\label{fig:4}
\end{centering}
\end{figure*}

 It can be seen from the numerical simulation that the architecture changes caused by topology or hyperparameter will lead to performance changes. Compared with the importance of the two, the influence of hyperparameter is at least not less than that of topology. About $6\%$ of accuracy improvement can be obtained by changing hyperparameters, while changing the topology can only obtain less than $2\%$ in the experiments. Based on the above analyses, when defining the PolSAR tailored search space, we focus on finding better combinations of hyperparameters under a specific topology in order to achieve cost-effective architecture search. Without loss of generality, the most commonly used cascaded CNN architecture \cite{7460942,7762055,complex,3dcnn,input3} is used as the topology to be searched, which has three convolution layers, two pooling layers and two fully connected layers. The search of DAS is carried out in this topology. Hyperparameters of the topology constitute the search space of DAS, i.e. the size of convolution kernels and the output depth of convolution operations. An intuitive diagram of the search space tailored for PolSAR classification is shown in Fig. \ref{fig:4}.
\par The height and width of convolution kernels, and the number of output channels are variables of the search space, which are noted as $S_h, S_w, C$. In this way, the total number of potential architectures $N_{total}$ in the proposed search space can be obtained:
\begin{equation}
N_{total}=\prod_{i=1}^lS_h^{i} \times S_w^{i} \times C^{i}\label{1}
\end{equation}
where $l$ means the number of layers of the used topology and $l=5$ in this work. It is worth noting that using such a fixed topology is to verify the validity of DAS as well as reduce the search complexity as much as possible. Searching hyperparameters within various kinds of architectures suitable for PolSAR classification are also supported by DAS.
\subsubsection{Continuous Relaxation}
\par After determining the appropriate search space, an efficient search strategy needs to be designed to implement NAS with a single GPU. Based on this requirement, continuous relaxation proposed by DARTS \cite{darts} is applied to our search space. Let $\mathcal{O}$ be a set of candidate operations, for the input $x$, continuous relaxation can be defined as follows:
\begin{equation}
\tilde{o}(x)=\sum_{o\in \mathcal{O}}\sigma(\mathcal{A}_o)o(x)\label{2}
\end{equation}
where $\mathcal{A}_o$ notes the architecture parameters assigned to operation $o(\cdot)$, and $\sigma(\cdot)$ means the weight of corresponding operation generated by an activation function. From the definition, continuous relaxation can be seen as using a mixed operation to replace the categorical choice of a particular operation. Softmax activation is used over architecture parameters to generate the weights of $o(\cdot)$ in \cite{darts}, and \eqref{2} specified to:
\begin{equation}
\begin{split}
\tilde{o}(x)& =\sum_{o\in \mathcal{O}}\sigma_{softmax}(\mathcal{A}_o)o(x) \\
& =\sum_{o\in \mathcal{O}} \frac{exp(\mathcal{A}_o)}{\sum_{o^{'}\in \mathcal{O}}exp(\mathcal{A}_{o^{'}})}o(x).\label{3}
\end{split}
\end{equation}
\par In this work, Sparsemax activation \cite{sparsemax} which can be seen as a truncated Softmax is used in DAS to increase the sparsity of architecture parameters. If the architecture parameters are sorted as $\mathcal{A}_{(1)}\geq \cdots\geq \mathcal{A}_{(K)}$, a set $k$ can be found with $k(\mathcal{A}):=\max\{k\in[K]|1+k\mathcal{A}_{(k)}>\sum_{j\leq k}\mathcal{A}_{(j)}\}$. Then the mixed operation of DAS can be written as:
\begin{equation}
\begin{split}
\tilde{o}(x) & =\sum_{o\in \mathcal{O}}\sigma_{sparsemax}(\mathcal{A}_o)o(x)\\
 & =\sum_{o\in \mathcal{O}} \max(0, \mathcal{A}_o-\tau(\mathcal{A}))o(x)\label{4}
\end{split}
\end{equation}
where $\tau(\cdot)$ is a threshold function defined as $\tau(\mathcal{A})=((\sum_{j\leq k(\mathcal{A})}\mathcal{A}_{(j)})-1)/k(\mathcal{A})$. The vector activated by Sparsemax tends to be a normalized sparse vector, which can accelerate the convergence of the search process.
\par Specifically, for the case of DAS, $o(x)$ represents convolution with different hyperparameter combination, and $\mathcal{A}=\{\alpha, \beta\}$ where $\alpha$ notes the architecture parameter of kernel size and $\beta$ for the one of output depth. If there are $n$ choices for kernel size and $m$ choices for output depth in a certain layer, \eqref{4} can be expressed as:
\begin{equation}
\begin{split}
\tilde{o}(x) & =\sum_{i=1}^m\sigma_{sparsemax}(\alpha_i)(\sum_{j=1}^n\sigma_{sparsemax}(\beta_j)(W_{i,j}\ast x))\\
 & =\sum_{i,j=1}^{n,m}\alpha_i^{'}\beta_j^{'}(W_{i,j}\ast x)\label{5}
\end{split}
\end{equation}
where the weights generated by Sparsemax activation are simplified as $\alpha^{'}, \beta^{'}$, and $W_{i,j}$ represents corresponding convolution matrix, $\ast$ means convolution operation. In this way, the discrete search space of DAS can be relaxed to be continuous, and the search of architectures can be transformed into the optimization of the architecture parameters $\alpha, \beta$.

\subsubsection{Linear Combination of Convolution}
\par Note that \eqref{2}-\eqref{5} need to calculate all possible operations and then weighted sum, which brings huge computational complexity. Especially for \eqref{5}, $nm$ times convolution is required, which is impractical when the product is very large. Therefore, linearity of convolution is utilized in DAS and multiple convolution calculations are avoided by using linear combination of convolutions instead of \eqref{5}, which can be expressed as:
\begin{equation}
\tilde{o}(x)=\sum_{i,j=1}^{n,m}(\alpha_i^{'}\beta_j^{'}W_{i,j})\ast x\label{6}
\end{equation}
\par To support the linear combination of convolutions, each kernel matrix must be the same size. However, different hyperparameter combinations lead to different size of kernel matrix. So transformations of each kernel matrix in both spatial and channel is needed. In DAS, we follow the transformation method of \cite{dnas}.
\begin{figure}[!h]
\begin{centering}
\includegraphics[width=6.5cm,height=4.2cm]{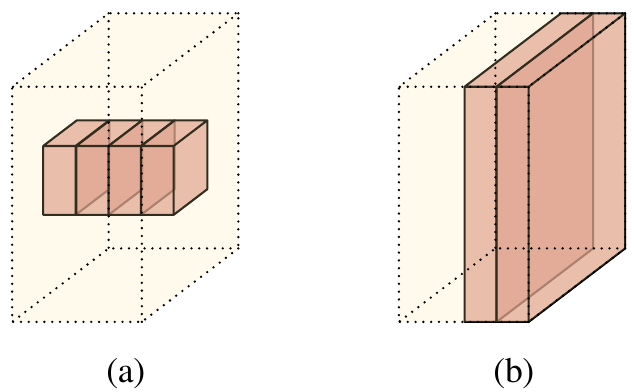}
\caption{Transformation of convolution kernels. The dark solid part represents the untransformed convolution kernel, and the light dotted part represents zero-padding. (a) The process of spatial transformation. (b) The process of channel-wise transformation.}\label{fig:7}
\end{centering}
\end{figure}
\par The process of spatial transformation can be seen from Fig. \ref{fig:7}a. Let $h_1\times w_1, h_2\times w_2, \cdots, h_n\times w_n$ denote all candidate kernel sizes. Based on the fact that convolution with the kernel of $h_i\times w_i$ is identical to the same convolution when its kernel is appropriately zero-padded to the size of $h_{max}\times w_{max}$, where $h_{max}=\max\{h_1, h_2,\cdots, h_n\}$ and $w_{max}=\max\{w_1, w_2,\cdots, w_n\}$. Thus, all kernel matrices are spatially padding to the size of $h_{max}\times w_{max}$ while keeping the original operation unchanged.
\par The process of channel-wise transformation can be seen from Fig. \ref{fig:7}b. A simple trick is used to realize the channel-wise transformation by adjusting the number of input channels instead of changing the one of output channels. In this way, all kernel matrices are channel-wise padding to the size of $c_{max}\times c_{out}$. In order to save computational resources, in this paper, $h_{max}$ and $w_{max}$ are set to $5$ for $l=1,2$, and $3$ for $l=3$. $h_{max}=w_{max}=1$ for the fully connected layers. In addition, $c_{max}$ is set to $64$ for $l=1,2,3$, and $1024$ for $l=4$. Similar ideas have emerged in some channel-wise pruning studies \cite{8416559} whose connotation is to use the statistical information of $i+1$th layer to guide the architecture search of $i$th layer.

\subsubsection{Two-Step One-Shot Architecture Search}
\par The computational difficulty is alleviated by the relaxation of search space and the linear combination of convolutions. In this part, we propose a two-stage one-shot architecture optimization method to further reduce the memory requirements as well as keep the search precision of DAS.
\par It can be seen that \eqref{6} contains $\sum_{i,j}h_i\times w_i\times c_j\times c_{out}$ weight parameters. Such a large number of trainable parameters not only causes the demand for memory, but also aggravates the difficulty of optimization. To handle this problem, $n\times m$ groups of weight parameters are merged into one with the size of $h_{max}\times w_{max}\times c_{max}\times c_{out}$ in \cite{dnas}. Although the number of parameters is reduced by parameter merging, untractable bias is introduced because all the branches share the parameter from the biggest kernel matrix. Instead of roughly relaxing $n\times m$ groups parameters to $1$, in DAS, a two-stage optimization for architecture parameters is used to relax $n\times m$ to $n + m$. Specifically, the optimization of $\alpha$ and $\beta$ is carried out in an alternating iterative way to avoid the multiplication of weight parameters caused by jointly optimization. When $\beta$ is fixed, \eqref{6} is reduced to the following expression:
\begin{equation}
\tilde{o}(x)=\sum_{i=1}^{n}(\alpha_i^{'}\beta^{'}W_{i})\ast x\label{6.1}
\end{equation}
and when $\alpha$ is fixed, it can be expressed as:
\begin{equation}
\tilde{o}(x)=\sum_{j=1}^{m}(\alpha^{'}\beta_j^{'}W_{j})\ast x.\label{6.2}
\end{equation}
\par Due to the intentionally diversified candidate architectures in DARTS, separate optimization of architecture and weight parameters are needed \cite{darc}, which leads to a bilevel optimization problem. In contrast, DAS has the PolSAR tailored search space with small diversity, so it is suitable to optimize the architecture and weight parameters on a single training set. We introduce the optimization problem of DAS as follows:
\begin{equation}
(\alpha^{*}, \beta^{*})=\mathop{\arg\min}\limits_{\alpha,\beta\in\mathcal{A}} \{L_{train}(\mathcal{A}, w) + \gamma R(\mathcal{A}) \}\label{7}
\end{equation}
which can be alternately optimized by:
\begin{equation}
\alpha^{*}=\mathop{\arg\min}\limits_{\alpha} \{L_{train}(\alpha, w, \beta_{fixed}) + \gamma R(\alpha) \}\label{8}
\end{equation}
and
\begin{equation}
\beta^{*}=\mathop{\arg\min}\limits_{\beta} \{L_{train}(\beta, w, \alpha^{*}) + \gamma R(\beta) \}\label{9}
\end{equation}
where $L_{train}(\cdot)$ notes cross-entropy loss on training set and $R(\cdot)$ is a regularization function weighted by $\gamma$. $L_1$ regularization is used in DAS to add sparsity to architecture parameters. The iterative search process of DAS for architecture parameters is outlined as Algorithm \ref{alg:1}. Once the optimal architecture parameters $\alpha, \beta$ are determined, the corresponding model can be constructed by keeping top-$k$ strongest operations of each optimal architecture parameter ($k=1$ in this work), and re-trained to achieve PolSAR classification.
\begin{algorithm}[h!]
\caption{Two-Step One-Shot Architecture Search for DAS}
\label{alg:1}
\begin{algorithmic}[1]
\STATE \textbf{Begin};
\STATE Prepare Training and validation sets $D=\{x_i,y_i\}_{i=1}^N$, $x_i\in R^{n\times n}$, positive integers $I_1,I_2$ and $B_1,B_2$, hyperparameter $\gamma$ and learning rate $\xi$.\\
\REQUIRE Searching for $\alpha^*$\\
\STATE Prepare $\beta_{fixed}$. Initialize weight parameters $w, b$ and architecture parameters $\alpha$. \\
\FOR{epoch in $I_1$:}
    \FOR{batch in $B_1$:}
        \STATE Forward propagation by \eqref{6.1} with trainable parameters of $w, \alpha$;\\
        \STATE Update $\alpha$ by gradient descent for \eqref{8};\\
    \ENDFOR
\ENDFOR
\ENSURE Searching for $\beta^*$;\\
\STATE Prepare $\alpha^*$. Initialize weight parameters $w, b$ and architecture parameters $\beta$. \\
\FOR{epoch in $I_2$:}
    \FOR{batch in $B_2$:}
        \STATE Forward propagation by \eqref{6.2} with trainable parameters of $w, \beta$;\\
        \STATE Update $\beta$ gradient descent for \eqref{9};\\
    \ENDFOR
\ENDFOR
\RETURN $\alpha^*, \beta^*$
\end{algorithmic}
\end{algorithm}

\subsection{Complex-Valued DAS}
The proposed DAS is extended to the complex domain for better performance (CV-DAS). Instead of only supporting real-valued calculations, all real operations are converted to complex operations in CV-DAS. In CV-CNNs \cite{complex,complex2,complex3}, the weight of filters should be complex numbers so the weight $w\in\mathbb{C}$ and bias $b\in\mathbb{C}$ of a layer can be expressed as $w=\Re(w)+j\Im(w)$ and $b=\Re(b)+j\Im(b)$. Complex-valued convolution can be defined as follows:
\begin{equation}
\begin{split}
x^{i+1}&=(\Re(W^{i})\ast\Re(x^{i})-\Im(W^{i})\ast\Im(x^{i}))\\
&+ j(\Re(W^{i})\ast\Im(x^{i})+\Im(W^{i})\ast\Re(x^{i})) \\
&+ \Re(B^{i})+j(\Im(B^i)) \label{10}
\end{split}
\end{equation}
where $x^{i},x^{i+1}$ denote the output feature map of $i-1$th and $i$th layer, $W^i$ and $B^i$ are the weight matrix and bias vector of $i$th layer. Complex ReLU ($\mathbb{C}$ReLU) activation \cite{complex4} is used in DAS for activate the complex-valued products of convolutions. It can be defined as:
\begin{equation}
\mathbb{C}ReLU(z)=\max(0,\Re(z))+j\max(0,\Im(z)).  \label{11}
\end{equation}
\par Complex-valued fully connected layer can be seen as a special complex-valued convolution layer with $1\times 1$ kernel size. A feature vector whose dimension is the number of target categories can be obtained after two times of fully connected layers, and its amplitude is used to make final predictions.
\par Other parts of DAS, including search space and search strategy, remain unchanged. Compared with real-valued DAS, CV-DAS can maintain the integrity of PolSAR data and mine the physical scattering mechanism hidden in the complex-valued covariance or coherency matrices.

\section{Experimental Results and Discussion}
\label{sec:4}
In this section, experiments used datasets and evaluation criteria are introduced firstly. Based on that, the performance of the proposed methods are demonstrated and analyzed under three widely used PolSAR benchmark datasets. The classification results of the whole map and testing accuracy under three different criteria are recorded and compared.

\subsection{Datasets Description}
We employ three widely-used PolSAR datasets in the experiments: AIRSAR San Francisco, AIRSAR Flevoland, ESAR Oberpfaffenhofen. Figs. \ref{fig:gt0}-\ref{fig:gt2} show their Pauli maps and ground truth maps respectively. Besides, Tables \ref{tab:tab0}-\ref{tab:tab2} show some details about the three benchmark datasets.
\subsubsection{AIRSAR San Francisco}
\begin{figure}[h]
\begin{centering}
\includegraphics[width=8.0cm]{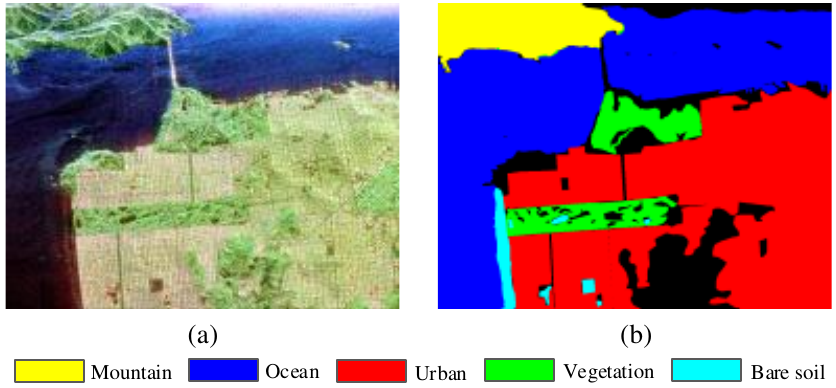}
\caption{AIRSAR San Francisco dataset and its color code. (a) Pauli RGB map. (b) Ground truth map.}\label{fig:gt0}
\end{centering}
\end{figure}

This PolSAR image has been widely used in literatures. It is an L-band, full polarimetric image of San Francisco, as shown in Fig. \ref{fig:gt0}. The size of this image is $900\times 1024$ and the spatial resolution is $0.6m\times1.6m$. Each pixel of this image can be classified into five categories, including mountain, ocean, urban, vegetation, and bare soil. The number of the labeled pixels can be seen in Table \ref{tab:tab0}.
\begin{table}[h]
\centering  
\caption{Number of Pixels in Each Class for Airsar San Francisco}
\renewcommand\arraystretch{1.25}
\begin{tabular}{p{54pt}|p{54pt}|p{54pt}c} 
\toprule
\multicolumn{3}{c}{\centering AIRSAR San Francisco} \\
\hline
\centering Class code &Name &Reference data\\ \hline  
\centering 1 &Mountain  &59035 \\         
\centering 2 &Ocean  &321583  \\
\centering 3 &Urban  &334418  \\
\centering 4 &Vegetation  &53509 \\         
\centering 5 &Bare soil  &13487  \\ \hline
\centering Total &- &782032  \\
\bottomrule
\end{tabular}
\label{tab:tab0}
\end{table}
\subsubsection{AIRSAR Flevoland}
\begin{figure}[h]
\begin{centering}
\includegraphics[width=8.0cm]{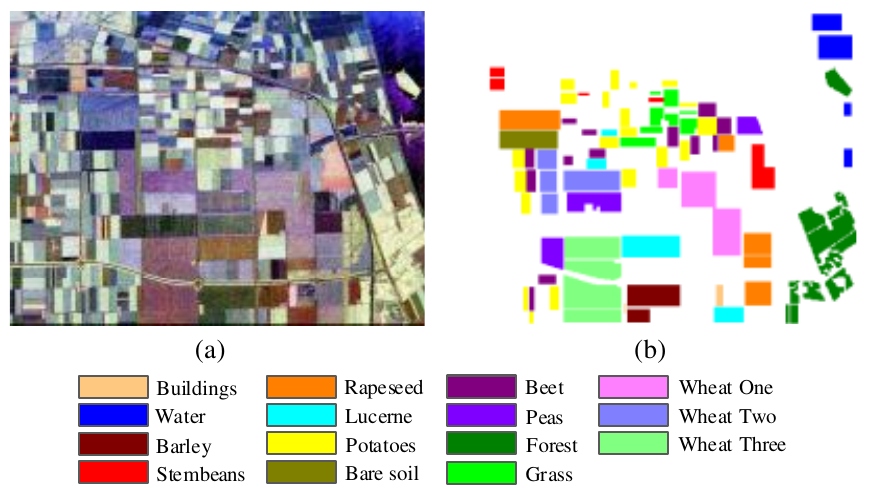}
\caption{AIRSAR Flevoland dataset and its color code. (a) Pauli RGB map. (b) Ground truth map.}\label{fig:gt1}
\end{centering}
\end{figure}

As shown in Fig. \ref{fig:gt1}, an L-band, full polarimetric image of the agricultural region of the Netherlands is obtained through NASA/Jet Propulsion Laboratory AIRSAR. The size of this image is $750\times1024$ and the spatial resolution is $0.6m\times1.6m$. There are $15$ kinds of ground objects including buildings, stembeans, rapeseed, beet, bare soil, forest, potatoes, peas, lucerne, barley, grasses, water and three kinds of wheat. The number of the labeled pixels can be seen in Table \ref{tab:tab1}.
\begin{table}[h]
\centering  
\caption{Number of Pixels in Each Class for Airsar Flevoland}
\renewcommand\arraystretch{1.25}
\begin{tabular}{p{54pt}|p{54pt}|p{54pt}c} 
\toprule
\multicolumn{3}{c}{\centering AIRSAR Flevoland} \\
\hline
\centering Class code &Name &Reference data\\ \hline  
\centering 1 &Buildings  &963 \\         
\centering 2 &Rapeseed  &17195  \\        
\centering 3 &Beet &11516\\
\centering 4 &Stembeans &6812  \\
\centering 5 &Peas &11394  \\
\centering 6 &Forest &20458 \\
\centering 7 &Lucerne &11411  \\
\centering 8 &Potatoes &19480  \\
\centering 9 &Bare soil &6116  \\
\centering 10 &Grass &8159  \\
\centering 11 &Barley &8046  \\
\centering 12 &Water &8824  \\
\centering 13 &Wheat one &16906  \\
\centering 14 &Wheat two &12728  \\
\centering 15 &Wheat three &24584  \\ \hline
\centering Total & - &184592  \\
\bottomrule
\end{tabular}
\label{tab:tab1}
\end{table}

\subsubsection{ESAR Oberpfaffenhofen}
\begin{figure}[h]
\begin{centering}
\includegraphics[width=9.0cm]{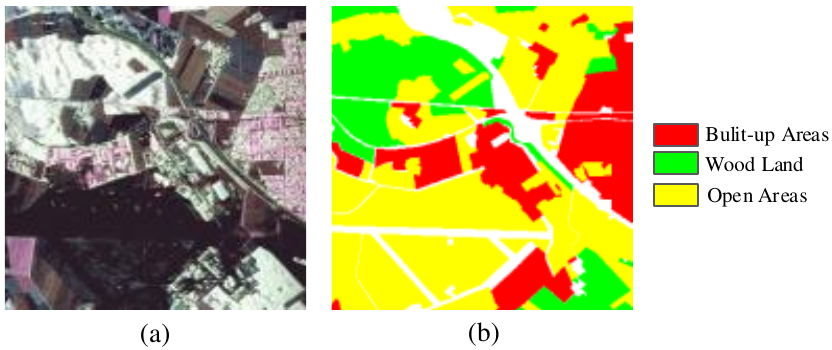}
\caption{ESAR Oberpfaffenhofen dataset and its color code. (a) Pauli RGB map. (b) Ground truth map.}\label{fig:gt2}
\end{centering}
\end{figure}

An L-band, full polarimetric image of Oberpfaffenhofen, Germany, $1200\times 1300$ scene size, are obtained through ESAR airborne platform. Its Pauli color-coded image and ground truth map can be seen in Fig. \ref{fig:gt2}. Each pixel in the map is divided into three categories: built-up areas, wood land and open areas, except for some unknown regions. The number of the labeled pixels can be seen in Table \ref{tab:tab2}.
\begin{table}[!t]
\centering  
\caption{Number of Pixels in Each Class for Esar Oberpfaffenhofen}
\renewcommand\arraystretch{1.25}
\begin{tabular}{p{54pt}|p{54pt}|p{54pt}c} 
\toprule
\multicolumn{3}{c}{\centering ESAR Oberpfaffenhofen} \\
\hline
\centering Class code &Name &Reference data\\ \hline 
\centering 1 &Built-up areas  &310829 \\         
\centering 2 &Wood land  &263238  \\
\centering 3 &Open areas  &733075  \\ \hline
\centering  Total&-  &1307142  \\
\bottomrule
\end{tabular}
\label{tab:tab2}
\end{table}

\subsection{Experimental Setting}
\subsubsection{Data Representations}
During the training and testing, PolSAR images are represented by the polarization coherence matrix T, which can be defined as:
\begin{equation}
\vec{k}=\frac{1}{\sqrt{2}}[S_{HH}+S_{VV}, S_{HH}-S_{VV}, 2S_{HV}]^T,\label{eqexp:1}
\end{equation}
\begin{equation}
[T]=\langle \vec{k}\vec{k}^{H}\rangle\label{eqexp:2}
\end{equation}
where $S_{PQ}(P, Q\in \{H, V\})$ represents the backscattering coefficient of the polarized electromagnetic wave in emitting $Q$ direction and receiving $P$ direction. $H$ and $V$ represent the horizontal and vertical polarization, respectively. According to the reciprocity theorem, the $S$ matrix satisfies $S_{HV}=S_{VH}$. Notice that the polarization coherence matrix $T$ is a Hermitian matrix, every element except the diagonal element is a complex number. Generally, we take the upper triangular elements $[T_{11}, T_{12}, T_{13}, T_{22}, T_{23}, T_{33}]$ as the input data. For real-valued models, complex-valued data are divide into real parts and imaginary parts as the input; For complex-valued models, original data are directly as the input. Benchmark PolSAR images are sliced into the size of $15\times 15$ image patches around the central pixel with the stride sliding windows of $1$ according to the ground truth map, as can be seen from Tables \ref{tab:tab0}-\ref{tab:tab2}, to generate the training, validation, and testing datasets.
\subsubsection{Comparing Methods and Parameter Settings}
The effectiveness of the proposed methods are proved by comparisons with some effective alternative methods. Three non deep learning based methods are chosen including Wishart classifier \cite{wishart}, support vector machines (SVM) \cite{svm}, and SVM with radial basis function kernel (RBF-SVM). Four deep learning based methods including CNN \cite{7762055}, complex-valued CNN (CV-CNN) \cite{complex}, and polarimetric-feature-driven CNN (SF-CNN) \cite{input1} are selected for comparison. The architectures searched by the proposed methods, i.e. DAS and CV-DAS, are noted as DAS-CNN and CVDAS-CNN for convenience.
\par Considering that the network structure obtained through the search is highly transferable \cite{8579005}, we do architecture search on a small dataset and then evaluate the obtained architectures in different benchmarks. In this work, the process of architecture search is carried out on Flevoland data, with $300$ training samples of each category and $200$ training epoch. Complete search process is repeated ten times to reduce the randomness of the searched architectures.
\par
\begin{figure}[!h]
\begin{minipage}{0.48\linewidth}
  \centerline{\includegraphics[width=5.0cm]{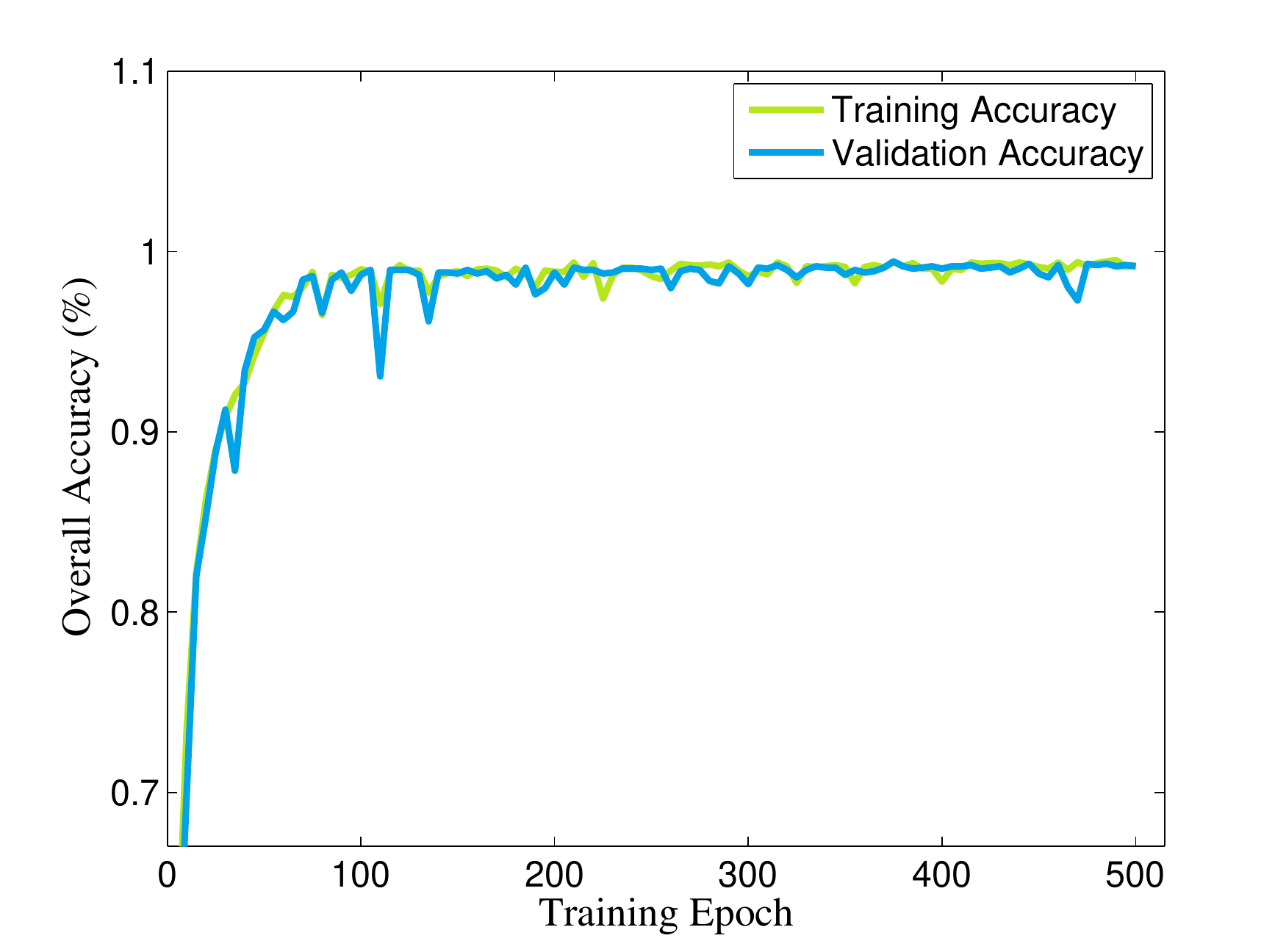}}
  \centerline{(a)}
\end{minipage}
\hfill
\begin{minipage}{0.48\linewidth}
  \centerline{\includegraphics[width=5.0cm]{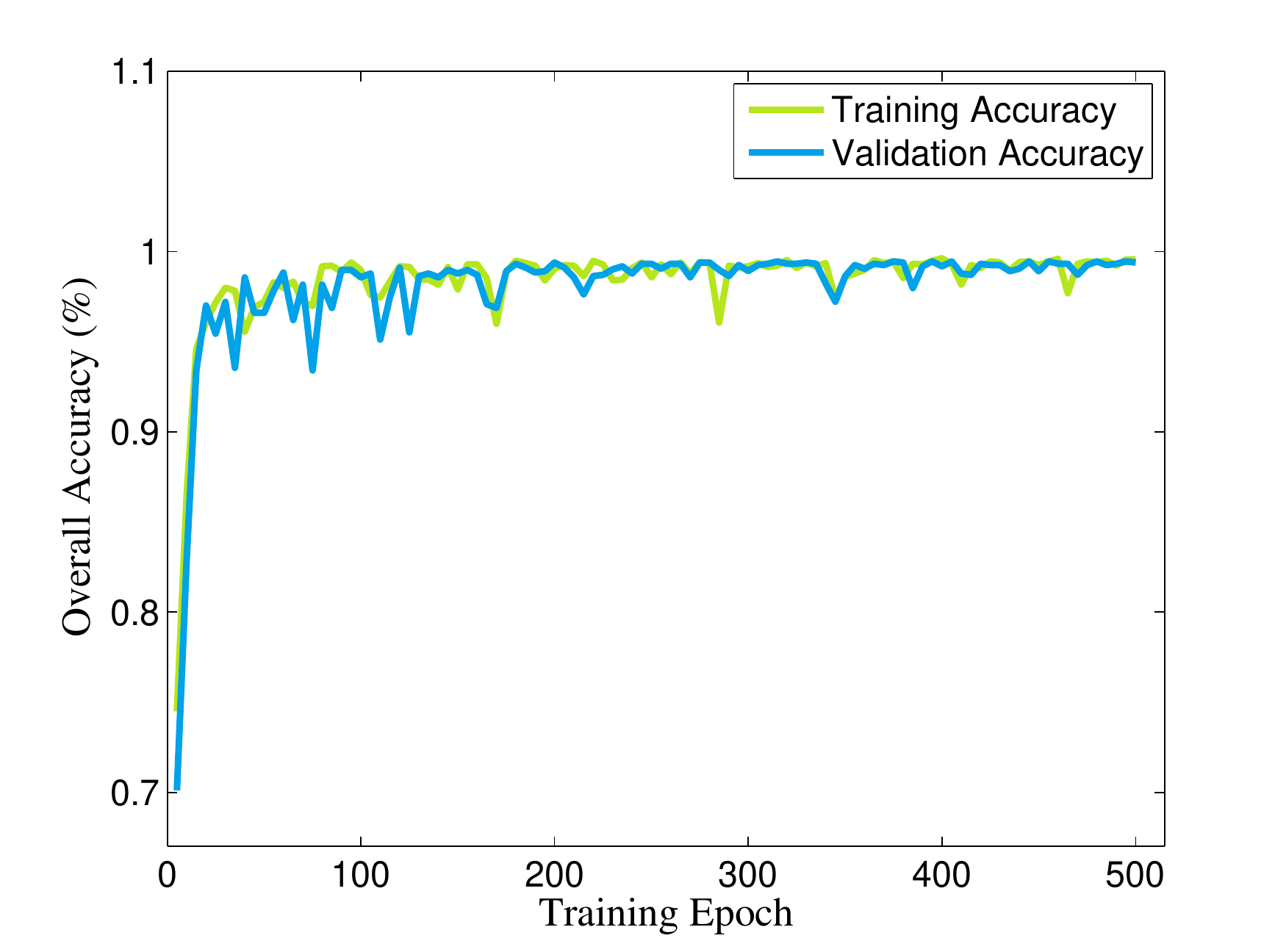}}
  \centerline{(b)}
\end{minipage}
\caption{Overall accuracy curves of the two methods on training and validation sets of San Francisco dataset respectively. (a) CNN. (b) DAS-CNN.}
\label{exp:epoch1}
\end{figure}

\begin{figure}[!h]
\begin{minipage}{0.48\linewidth}
  \centerline{\includegraphics[width=5.0cm]{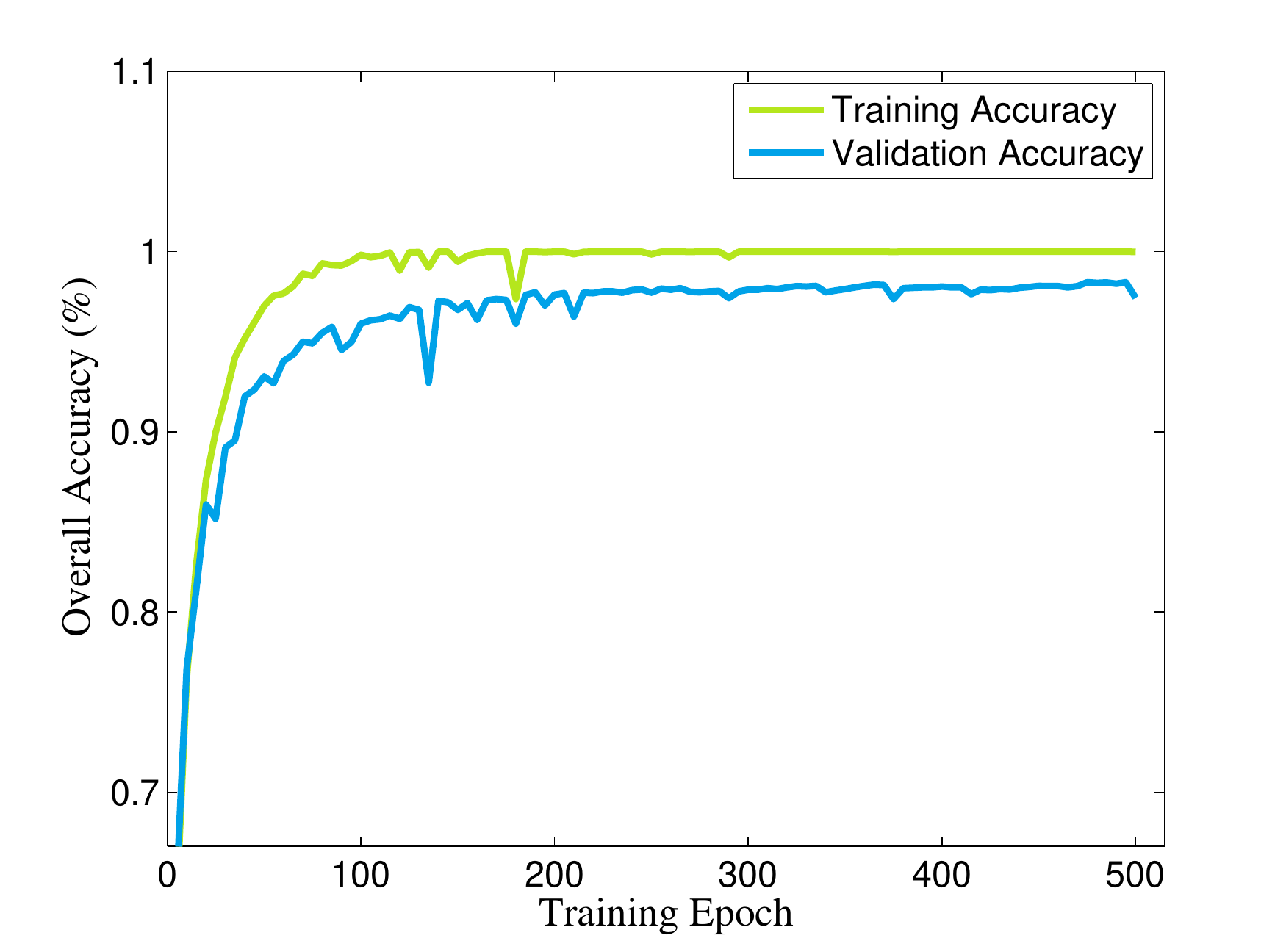}}
  \centerline{(a)}
\end{minipage}
\hfill
\begin{minipage}{0.48\linewidth}
  \centerline{\includegraphics[width=5.0cm]{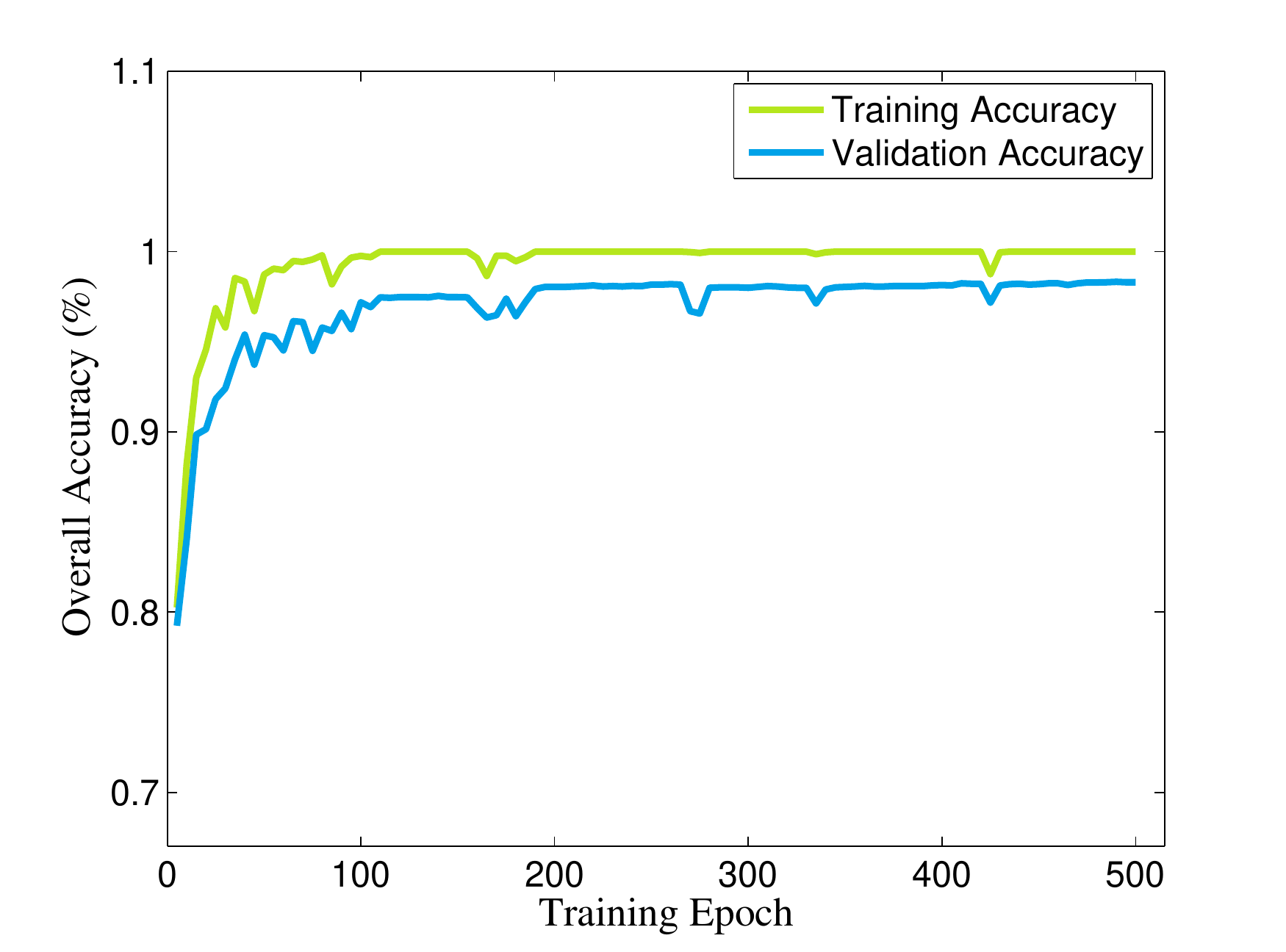}}
  \centerline{(b)}
\end{minipage}
\caption{Overall accuracy curves of the two methods on training and validation sets of Flevoland dataset respectively. (a) CNN. (b) DAS-CNN.}
\label{exp:epoch2}
\end{figure}

\begin{figure}[!h]
\begin{minipage}{0.48\linewidth}
  \centerline{\includegraphics[width=5.0cm]{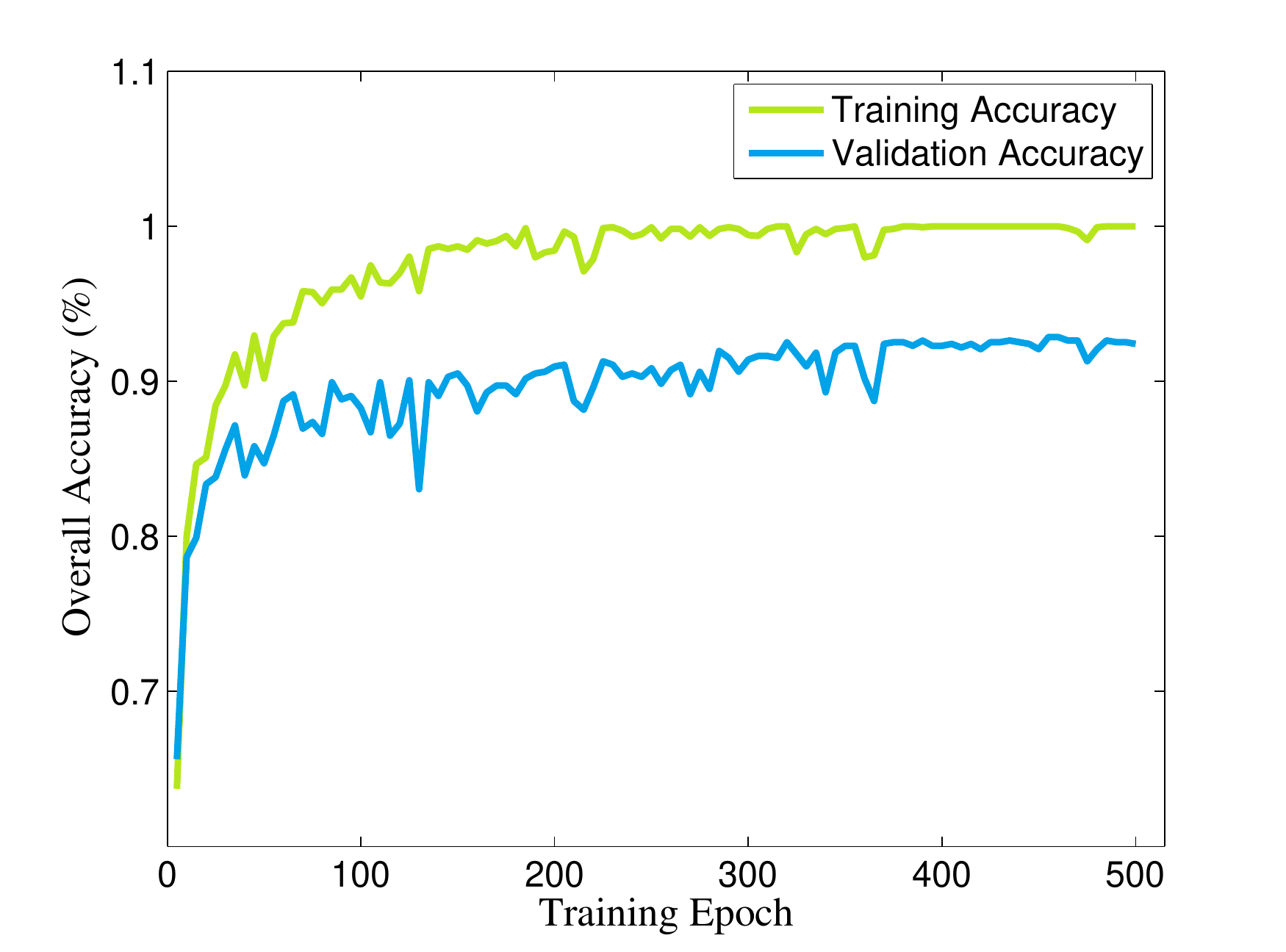}}
  \centerline{(a)}
\end{minipage}
\hfill
\begin{minipage}{0.48\linewidth}
  \centerline{\includegraphics[width=5.0cm]{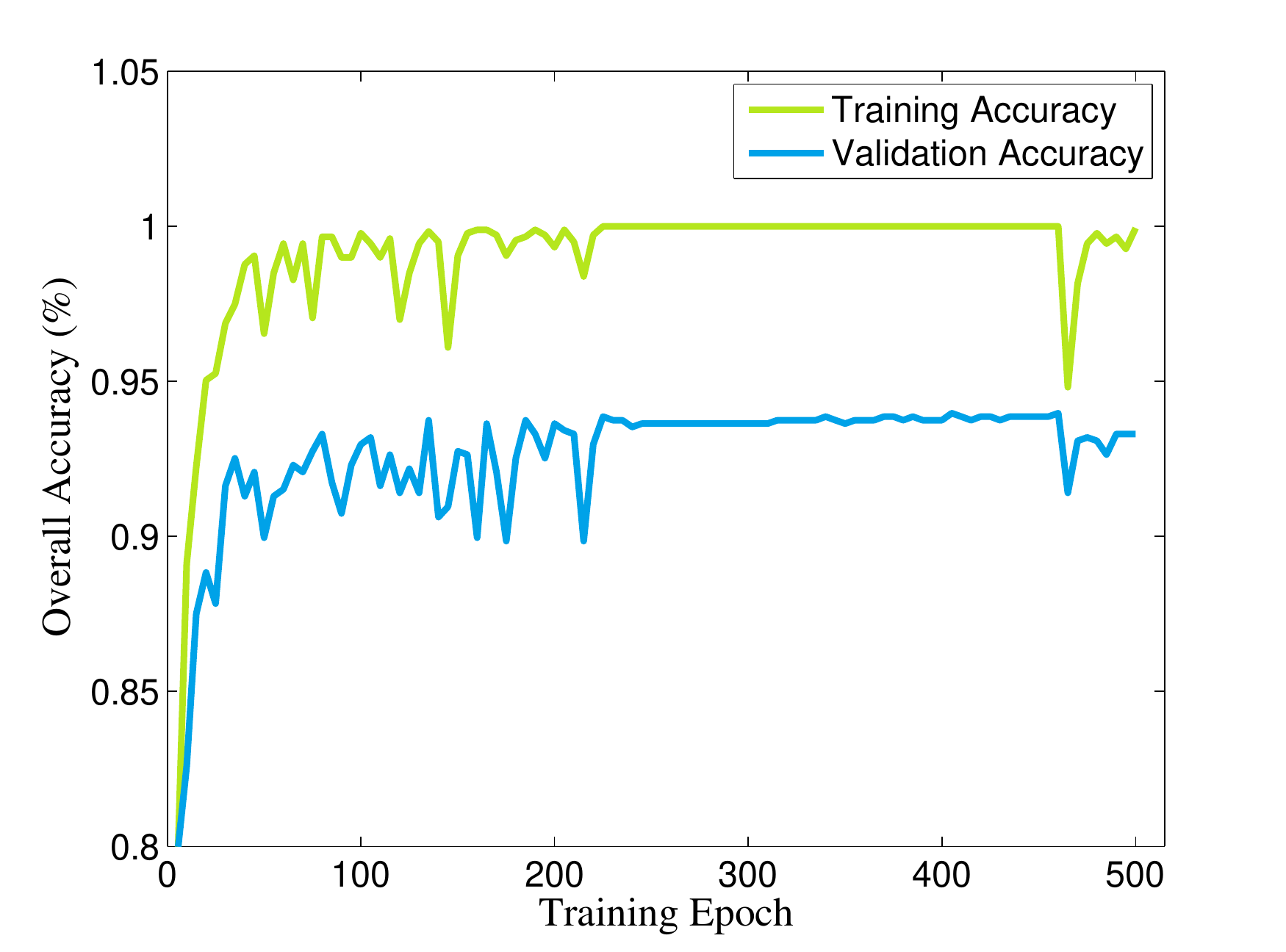}}
  \centerline{(b)}
\end{minipage}
\caption{Overall accuracy curves of the two methods on training and validation sets of Oberpfaffenhofen dataset respectively. (a) CNN. (b) DAS-CNN.}
\label{exp:epoch3}
\end{figure}

 After the architecture search, the model retraining is needed to achieve the final classification. The number of training epoch is an important hyperparameter for the training of all involved methods, which determines whether weights of the network converge or not. In order to find a suitable value of epoch on every datasets, experiments are carried out and the results are shown in Figs. \ref{exp:epoch1}-\ref{exp:epoch3}.
\par From the experimental results one can see that the error rate of real-valued CNN and DAS-CNN on training and validation sets tend to be stable within $500$ iterations. Therefore, the value of training epoch is set to $500$ in all experiments and a small validation set is used to select the best model from many iterations. Some other setting as follows: After each convolution layer, a ReLU activation layer is added. In order to speed up the optimization of objective function and obtain better approximate solution, stochastic gradient descent with adaptive moment estimation \cite{adam} are used with the learning rate of $10^{-4}$. Number of mini-batch is set to $64$ during the training. Deep learning toolbox \cite{tensorflow} is utilized to minimize the difficulty of algorithm implementation.
\subsubsection{Evaluation Criteria}
To evaluating the performance of the algorithms mentioned in this paper, overall accuracy (OA), average accuracy (AA), and kappa coefficient (Kappa) are chosen as criteria, which can be defined as follows
\begin{equation}
OA=\frac{\sum_{i=1}^c M_i}{\sum_{i=1}^c N_i}, \; AA=\frac{1}{c}\sum_{i=1}^c\frac{M_i}{N_i}
\label{eqexp:3}
\end{equation}
where $c$ is the number of categories. $M_i,N_i$ denote the number of $i$th category correctly classified samples and the number of the $i$th category labeled samples respectively.
\begin{equation}
Kappa=\frac{OA-P}{1-P}, \; with \; P=\frac{1}{N^2}\sum_{i=1}^cH(i,:)H(:,i)
\label{eqexp:4}
\end{equation}
where $N$ is the number of testing samples and $H$ denotes the classification confusion matrix.

\subsection{Experimental Results}
\begin{figure*}[!t]
\begin{centering}
\includegraphics[width=\textwidth]{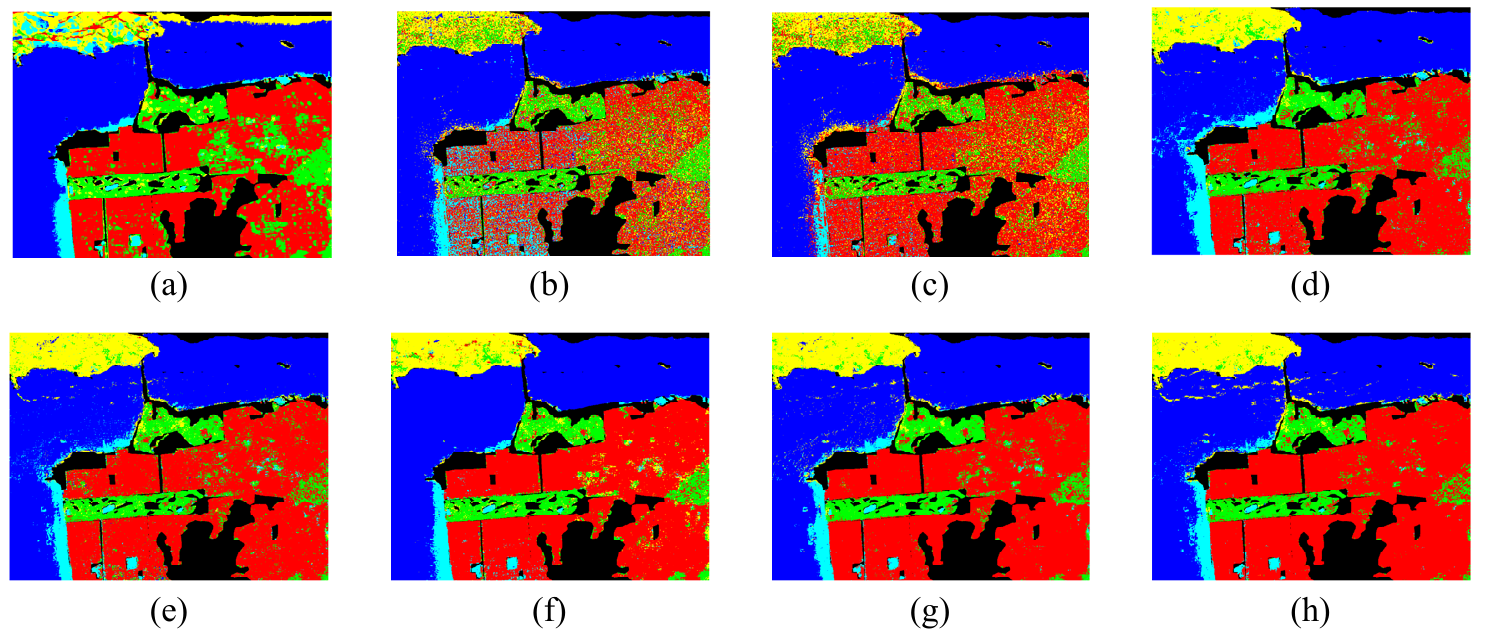} 
\caption{Classification results of whole map on AIRSAR San Francisco data with different methods. (a) Result of Wishart. (b) Result of SVM. (c) Result of RBF-SVM. (d) Result of CNN. (e) Result of CV-CNN. (f) Result of SF-CNN. (g) Result of DAS-CNN. (h) Result of CVDAS-CNN.}\label{fig:san}
\end{centering}
\end{figure*}

\begin{figure*}[!t]
\begin{centering}
\includegraphics[width=\textwidth]{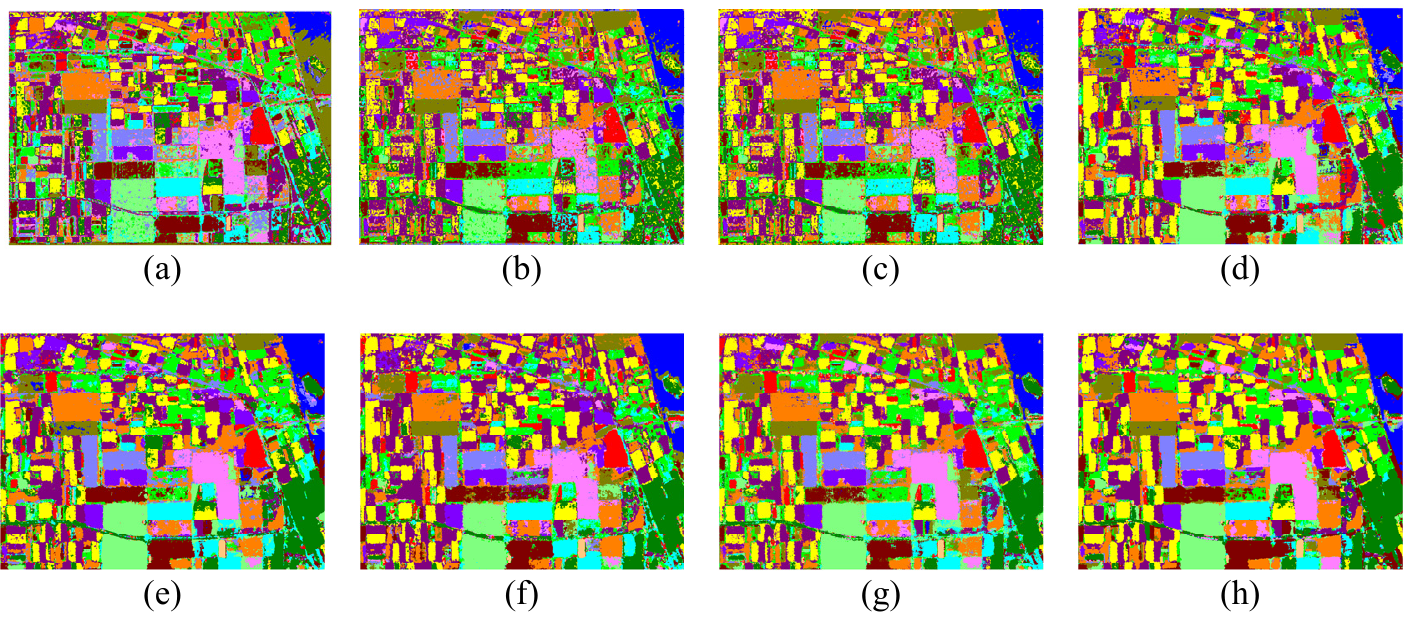} 
\caption{Classification results of whole map on AIRSAR Flevoland data with different methods. (a) Result of Wishart. (b) Result of SVM. (c) Result of RBF-SVM. (d) Result of CNN. (e) Result of CV-CNN. (f) Result of SF-CNN. (g) Result of DAS-CNN. (h) Result of CVDAS-CNN.}\label{fig:fle}
\end{centering}
\end{figure*}

\begin{figure*}[!t]
\begin{centering}
\includegraphics[width=16.0cm]{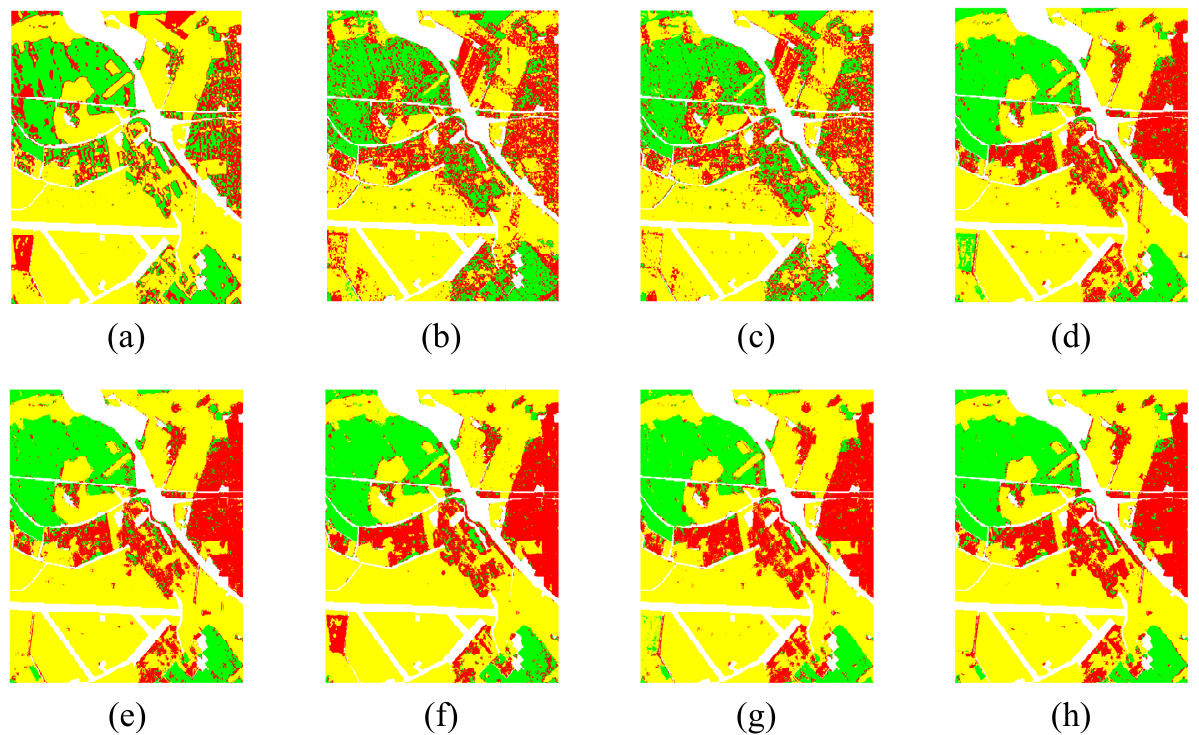} 
\caption{Classification results of whole map on ESAR Oberpfaffenhofen data with different methods. (a) Result of Wishart. (b) Result of SVM. (c) Result of RBF-SVM. (d) Result of CNN. (e) Result of CV-CNN. (f) Result of SF-CNN. (g) Result of DAS-CNN. (h) Result of CVDAS-CNN.}\label{fig:esar}
\end{centering}
\end{figure*}

The experimental results from testing sets on involved three PolSAR benchmarks are shown in Tables \ref{tab:san}-\ref{tab:esar}. The whole map classification results are shown in Figs \ref{fig:san}-\ref{fig:esar}. Generally speaking, deep learning based methods show more powerful classification ability than non deep learning methods. Moreover, two improved version, i.e. CV-CNN and SF-CNN have better performance than the ordinary CNN. Although no attention has been paid to the neural design, architectures searched by the proposed DAS and CV-DAS emerge the best generalization performance in three benchmark datasets. The experimental results on each dataset are analyzed as follows.
\begin{figure*}[!t]
\begin{minipage}{0.32\linewidth}
  \centerline{\includegraphics[width=5.5cm,height=5.0cm]{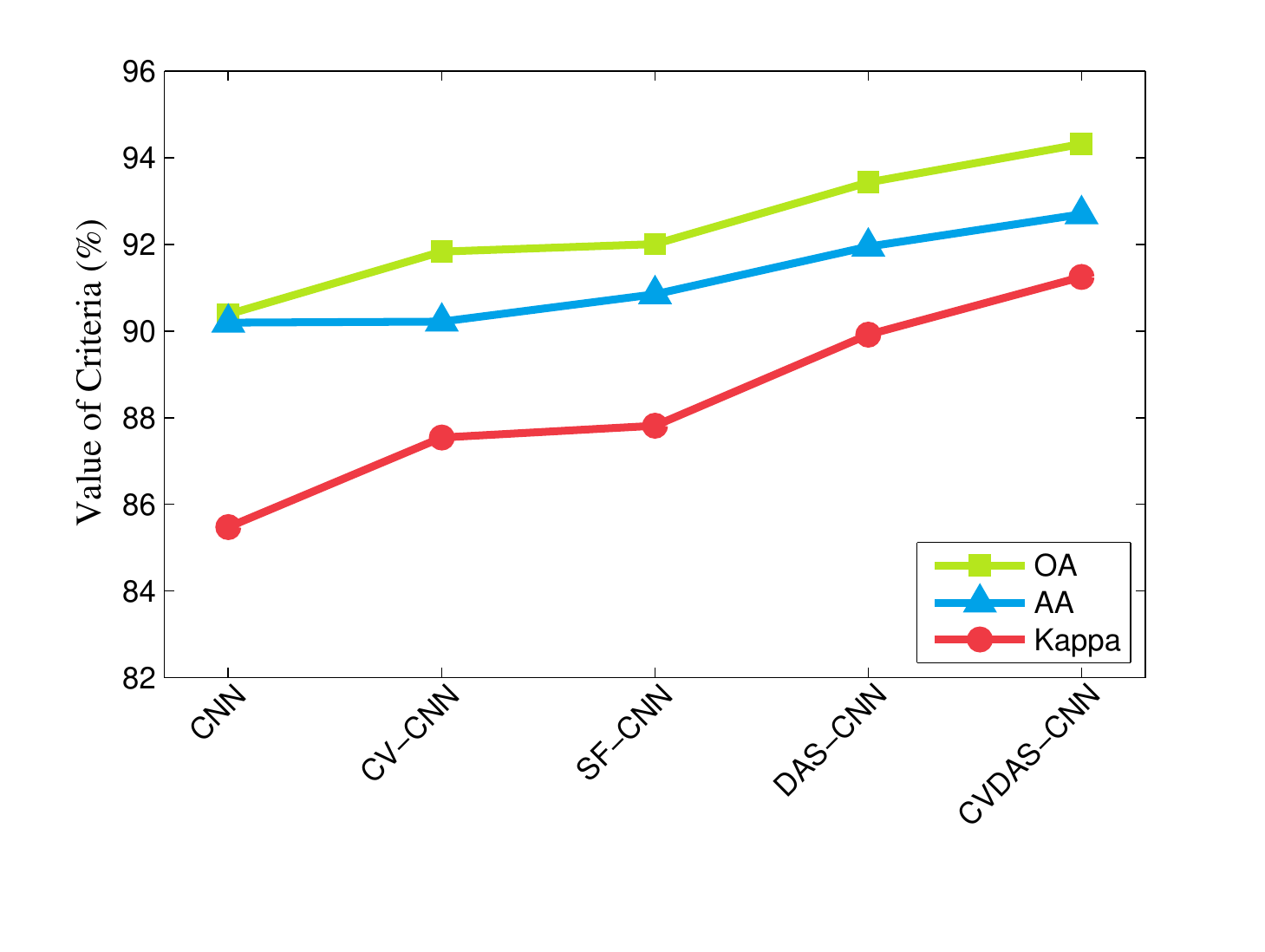}}
  \centerline{(a)}
\end{minipage}
\hfill
\begin{minipage}{0.32\linewidth}
  \centerline{\includegraphics[width=5.5cm,height=5.0cm]{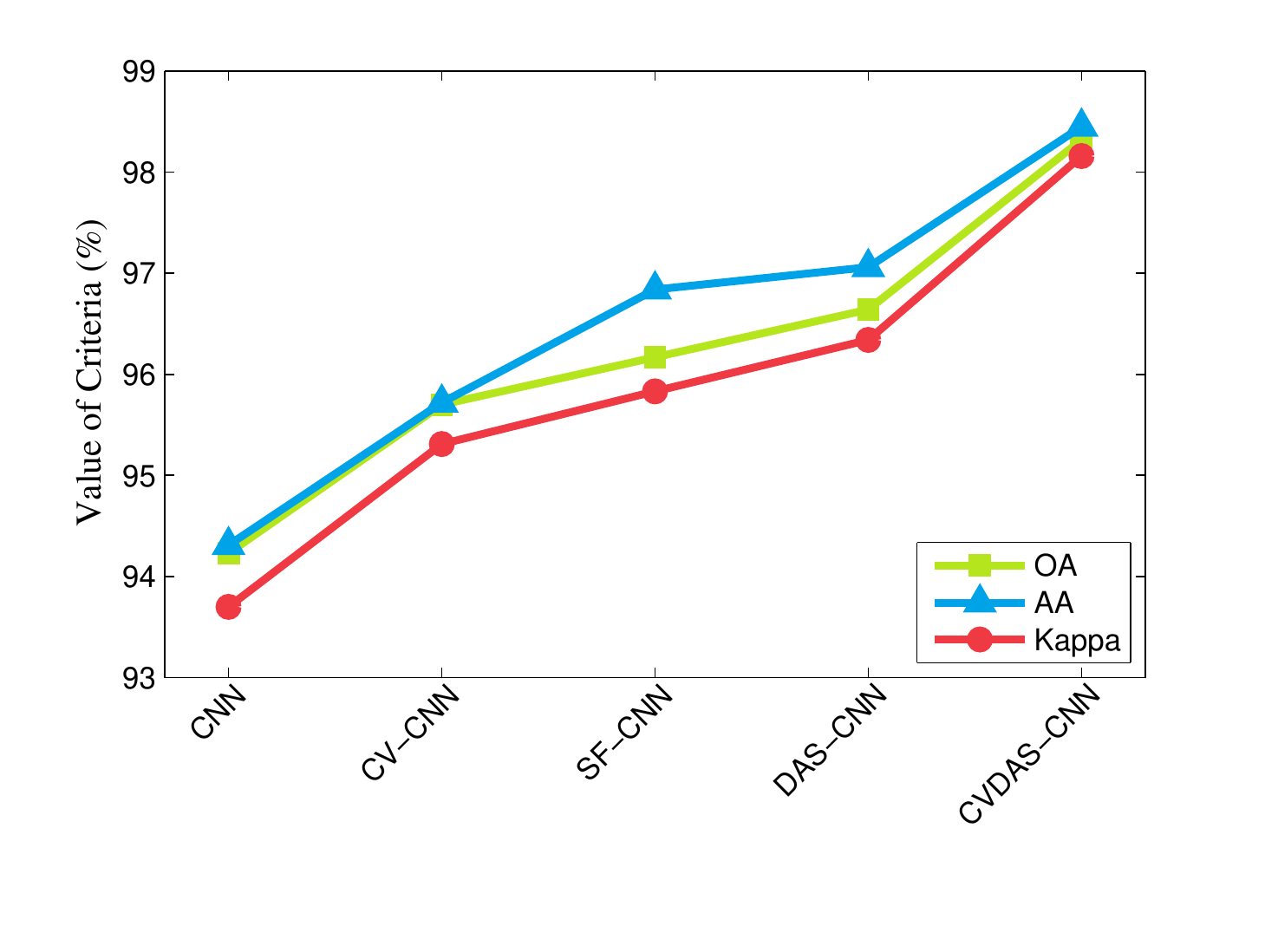}}
  \centerline{(b)}
\end{minipage}
\hfill
\begin{minipage}{0.32\linewidth}
  \centerline{\includegraphics[width=5.5cm,height=5.0cm]{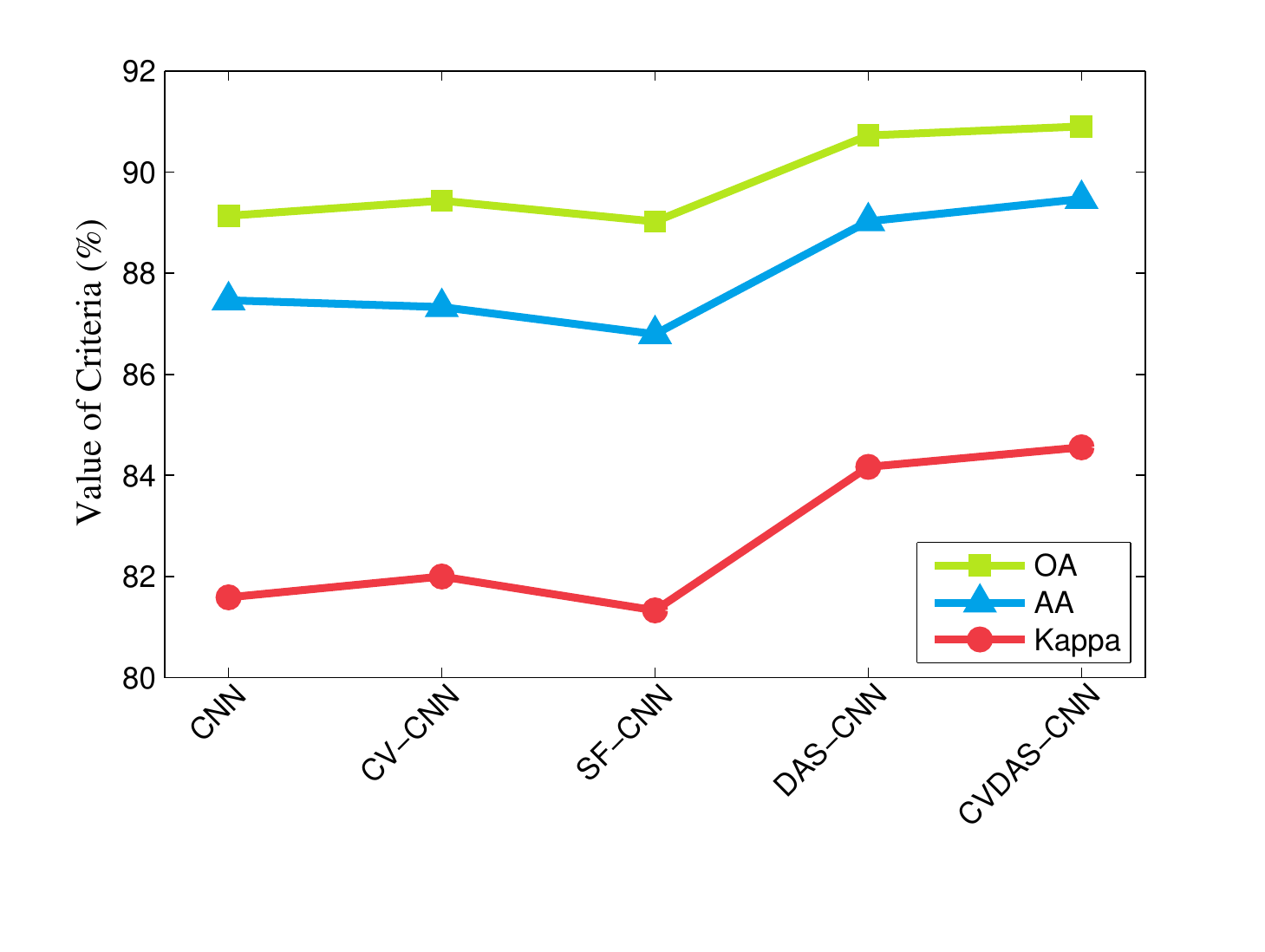}}
  \centerline{(c)}
\end{minipage}
\caption{Performance comparison of involved deep learning based methods on three benchmarks. (a) Result on San Francisco. (b) Result on Flevoland. (c) Result on Oberpfaffenhofen.}
\label{exp:compare}
\end{figure*}
\par Classification results of the whole map on San Francisco data can be seen from Fig. \ref{fig:san}. Architectures obtained by the proposed search methods, especially CVDAS-CNN, achieve higher completeness of the terrains in the classification maps. It can be seen from the results that the proposed methods have better recognition accuracy than the comparing methods for the identification of urban and vegetation. The proposed methods also avoid the situation that urban are wrongly divided into mountain in SF-CNN. This observation proves that the proposed methods can find better CNN architectures to achieve good classification results on San Francisco dataset. The experimental results on testing set are shown in Table \ref{tab:san}, it can be seen that the proposed methods achieve the best and the second results on three criteria, respectively. $3.05\%$ and $4.45\%$ increase of OA and Kappa are accomplished through DAS for real-valued CNN and $2.49\%$, $3.71\%$ for complex-valued CNN. This can verify the correctness of the theoretical analysis from the experimental point of view.
\par With the addition of DAS, the performance of CNN has been improved in all categories except for the category bare soil. This shows that DAS has good adaptability in the real-valued topology in experiments. After the search of CV-DAS, the architecture has a little degeneration in ocean and bare soil categories, and improved in other categories, especially in the recognition of mountain category.
\begin{table*}[!t]
\centering  
\caption{Classification Result Comparisons ($\%$) for Airsar San Francisco Dataset.}
\renewcommand\arraystretch{1.25}
\begin{tabular}{C{1.5cm}C{1.2cm}C{1.2cm}C{1.3cm}C{1.2cm}C{1.2cm}C{1.3cm}C{1.3cm}C{1.65cm}}  
\toprule
Method   &Wishart&SVM&RBF-SVM&CNN&CV-CNN&SF-CNN&DAS-CNN&CVDAS-CNN \\
\midrule
Mountain&56.25 &64.45 &67.40 &90.99 &88.19 &87.50 &92.43 &\textbf{95.12} \\
Ocean&91.67 &\textbf{98.19} &97.97 &93.24 &95.61 &96.79 &95.62 &94.66 \\
Urban&76.24 &66.49 &82.00 &88.57 &90.05 &89.34 &92.77 &\textbf{95.28} \\
Vegetation&80.59 &68.34 &64.80 &82.47 &83.83 &83.49 &85.38 &\textbf{85.68} \\
Bare soil&86.69 &83.09 &62.59 &95.70 &93.40 &\textbf{97.11} &93.55 &92.73 \\
\midrule
OA &81.49 &75.89 &81.34 &90.38 &91.83 &92.01 &93.43 &\textbf{94.32} \\
AA &78.29 &76.11 &74.95 &90.19 &90.22 &90.85 &91.95 &\textbf{92.69} \\
Kappa &72.87 &67.63 &74.03 &85.47 &87.54 &87.82 &89.92 &\textbf{91.25} \\
\bottomrule
\end{tabular}
\label{tab:san}
\end{table*}
\par The whole map classification results on Flevoland data are presented in Fig. \ref{fig:fle}, from which one can see that the classification results of the searched architectures are clearer and more close to the ground truth map. Table \ref{tab:fle} shows the experimental results of each algorithm on Flevoland dataset. The proposed two methods achieve the best and the second performance. With the help of DAS and CV-DAS, the recognition accuracy of real-valued and complex-valued architectures in almost every category has been improved, especially for the categories of rapeseed, bare soil and grass. From the global point of view, the architecture searched by CV-DAS gets $98.32\%$ OA, which is much higher than the results of other comparing methods. The process of search improves $2.41\%$ OA and $2.64\%$ Kappa for real-valued CNN, $2.62\%$ OA and $2.85\%$ Kappa for complex-valued CNN. This further proves that the proposed two search methods can improve the performance of a specific architecture in a plug-and-play manner.
\begin{table*}[!t]
\centering  
\caption{Classification Result Comparisons ($\%$) for Airsar Flevoland Dataset.}
\renewcommand\arraystretch{1.25}
\begin{tabular}{C{1.5cm}C{1.2cm}C{1.2cm}C{1.3cm}C{1.2cm}C{1.2cm}C{1.3cm}C{1.3cm}C{1.65cm}}  
\toprule
Method   &Wishart&SVM&RBF-SVM&CNN&CV-CNN&SF-CNN&DAS-CNN&CVDAS-CNN\\
\midrule
Buildings&68.02 &88.47 &87.54 &\textbf{100.00} &99.58 &\textbf{100.00} &\textbf{100.00} &\textbf{100.00}\\
Rapeseed&50.16 &81.38 &85.04 &80.77 &92.21 &93.79 &89.82 &\textbf{96.97}\\
Beet&91.86 &90.11 &90.91 &94.54 &95.33 &\textbf{98.82} &95.15 &96.18\\
Stembeans&93.04 &85.26 &88.21 &98.96 &98.22 &99.47 &\textbf{99.90} &99.44\\
Peas&87.06 &94.30 &94.38 &98.21 &98.44 &97.54 &98.10 &\textbf{98.97}\\
Forest&72.94 &82.22 &85.19 &97.85 &99.45 &96.96 &98.39 &\textbf{98.87}\\
Lucerne&90.82 &87.91 &89.10 &93.82 &97.68 &\textbf{98.04} &96.29 &96.34\\
Potatoes&58.83 &74.14 &81.50 &95.30 &87.33 &94.36 &95.00 &\textbf{98.61}\\
Bare soil&94.46 &95.14 &94.72 &90.79 &83.65 &99.13 &\textbf{99.33} &98.14\\
Grass&65.36 &70.52 &66.01 &82.14 &91.85 &92.82 &95.21 &\textbf{97.24}\\
Barley&94.24 &95.45 &94.95 &96.91 &99.58 &97.92 &96.96 &\textbf{99.95}\\
Water&52.95 &97.08 &97.67 &99.77 &99.99 &99.90 &\textbf{100.00} &\textbf{100.00}\\
Wheat one&89.76 &70.16 &79.11 &96.43 &96.84 &96.00 &98.38 &\textbf{98.92}\\
Wheat two&84.41 &82.79 &78.57 &90.26 &96.67 &92.51 &94.72 &\textbf{98.62}\\
Wheat three&82.44 &78.66 &86.73 &\textbf{98.96} &\textbf{98.96} &95.29 &98.69 &98.49\\
\midrule
OA&77.15 &82.69 &85.77 &94.23 &95.70 &96.17 &96.64 &\textbf{98.32}\\
AA&78.42 &84.91 &86.64 &94.31 &95.72 &96.84 &97.06 &\textbf{98.45}\\
Kappa&75.15 &81.18 &84.50 &93.70 &95.31 &95.83 &96.34 &\textbf{98.16} \\
\bottomrule
\end{tabular}
\label{tab:fle}
\end{table*}

\par Fig. \ref{fig:esar} reports the whole map classification results on Oberpfaffenhofen dataset. It can be seen from the results that there is a certain deviation in the recognition of bulit-up areas (red area) and wood land (green area) by ordinary architectures, and the addition of DAS and CV-DAS correct this deviation to a certain extent. Table \ref{tab:esar} shows the results of involved methods on testing set, from which one can see that the addition of the proposed search methods has brought certain improvement to the three evaluation criteria of the comparing methods. DAS is capable of getting accuracy increments of $1.58\%$, $1.57\%$, and $2.58\%$ for OA, AA, and Kappa coefficient, respectively. For CV-DAS, the value of increments changes to $1.47\%$, $2.14\%$, and $2.55\%$. These results indicate that the automatically optimized architectures obtained by the proposed search methods do have superior performance compared to the hand-crafted ones.
\begin{table*}[!t]
\centering  
\caption{Classification Result Comparisons ($\%$) for Esar Flevoland Dataset.}
\renewcommand\arraystretch{1.25}
\begin{tabular}{C{1.8cm}C{1.2cm}C{1.2cm}C{1.3cm}C{1.2cm}C{1.2cm}C{1.3cm}C{1.3cm}C{1.65cm}}  
\toprule
Method   &Wishart&SVM&RBF-SVM&CNN&CV-CNN&SF-CNN&DAS-CNN&CVDAS-CNN\\
\midrule
Built-up areas&45.19&55.35&47.20&77.85&77.02&78.09&\textbf{81.55}&80.48 \\
Wood land&74.87&79.38&90.07&91.44&90.75&88.43&91.03&\textbf{93.55} \\
Open areas&\textbf{94.49}&89.48&91.35&93.10&94.23&93.87&94.51&94.38 \\
\midrule							
OA&78.80&77.83&79.15&89.14&89.44&89.02&90.72&\textbf{90.91} \\
AA&71.52&74.07&76.21&87.46&87.33&86.80&89.03&\textbf{89.47} \\
Kappa&63.39&65.00&66.98&81.59&82.00&81.33&84.17&\textbf{84.55} \\
\bottomrule
\end{tabular}
\label{tab:esar}
\end{table*}
\par Comparisons between five CNNs-based methods are shown in Fig. \ref{exp:compare}, from which one can see that the proposed two methods have better performance on each benchmark dataset for the three evaluation criteria. It is worth noting that the architectures of DAS-CNN and CVDAS-CNN used in three groups of experiments are obtained by searching on Flevoland data. From the experimental results it can be seen that the transfer of the two architectures to the other two dataset still shows quite good performance, which proves that the proposed search methods have good robustness.

\subsection{Discussion}
The above experimental results show that the architectures obtained by the proposed DAS and CV-DAS achieved the state of the art performance. A global analysis and discussion are given after experimental simulations.
\par Firstly, thanks to the continuous relaxed search space, the proposed methods can be optimized by SGD to do one-shot architecture search, which brings about a very fast search speed. In our experimental environment, it takes only a few hours to implement once architecture search. This is beneficial for others to repeat and follow-up our work. Not only that, the architectures obtained by searching have a strong diversity compared with the current widely used ones in PolSAR area. Convolution kernels with $3 \times 3$ spatial size have been the choice of mostly existing studies, because its performance in optical image processing tasks is quite good. However, in our experiments, the size of $3 \times 3$ can not show an overwhelming advantage. $2 \times 1$, $5 \times 5$ and $1 \times 1$ convolutions are used in DAS-CNN and $3 \times 1$, $1 \times 1$ and $1 \times 1$ ones are used in CVDAS-CNN. This result also shows the difference between PolSAR image classification and optical image classification. We suspect that the reason for this phenomenon may be that image slices require a smaller convolution kernel to keep more details. The selected number of nodes of fully connected layers also show good diversity, but the trend of change is roughly predictable, that is, performance increases as the number of nodes adds. Furthermore, the improved optimization method of DAS avoids the bilevel optimization problem that classical gradient based NAS methods need to face. In our test, this improvement indeed improves the stability of the search process. Last but not least, the proposed search method is highly flexible and can be extended to any network architectures in any tasks to optimize their hyperparameters. Therefore, the proposed DAS and CV-DAS are not just searching suitable architectures for PolSAR image classification from scratch, they also can be used as plug-and-play complementary components to improve the performance of existing methods.

\section{Conclusion}\label{sec:5}
In this paper, the task of PolSAR image classification is promoted with the help of NAS for the first time. On basis of the gradient-based NAS, a tailored search method for PolSAR classification is proposed, which fully considers the form of PolSAR data and the task characteristics of classification. In the proposed DAS, search space and search strategy are specially improved in order to automatically design suitable architectures for PolSAR classification. Compared with the ordinary paradigm of designing PolSAR classification used architectures manually, the proposed methods are quite cost-effective. Furthermore, the complex-valued version of DAS is introduced to fit the data format of PolSAR images. It is shown that searched architectures have more powerful performance and better interpretability than hand-crafted ones.
\par Experiments on three widely used PolSAR benchmark datasets show that the architectures obtained by the proposed search methods have certain advantages over ordinary hand-crafted ones, which is reflected in the higher accuracy of the proposed methods in testing sets, and more close to the ground truth map in the whole map classification. The main reason why the proposed method has more powerful performance is that most of the PolSAR classification used CNN architectures are only one path in the DAS's search space. The proposal of DAS indicates the future research direction of deep learning technique in PolSAR image classification. For our future works, better input forms of PolSAR data, high-precision NAS methods and their application to PolSAR semantic segmentation are all the issues we are considering.


%

%
%
%
%
%
%
%

\bibliographystyle{IEEEtran}
\bibliography{IEEEabrv,refference}

\begin{IEEEbiography}[{\includegraphics[width=1in,height=1.25in,clip,keepaspectratio]{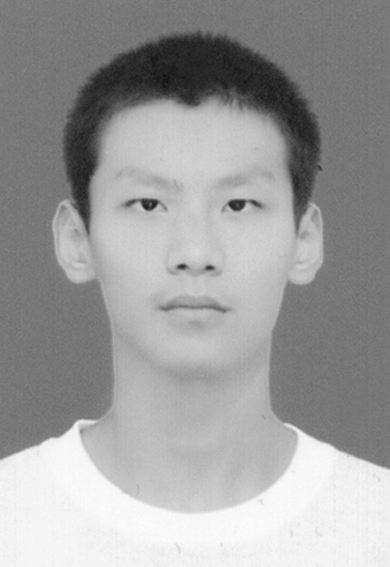}}]{Hongwei Dong} received the M.S. degree from College of Science, China Agricultural University, Beijing, China, in 2018.  He is currently pursuing the Ph.D. degree in Harbin Institute of Technology, Harbin, China. His current research interests include applied mathematics, optimization methods, machine learning and neural architecture search.
\end{IEEEbiography}
\begin{IEEEbiography}[{\includegraphics[width=1in,height=1.25in,clip,keepaspectratio]{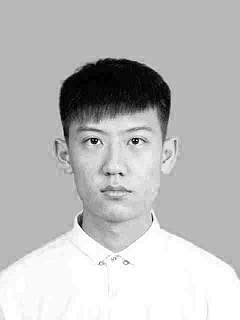}}]{Siyu Zhang} received the B.S. degree from Electronic and Information Engineering, Dalian Maritime University, Dalian, China, in 2019. He is currently pursuing the M.S. degree with Information and Communication Engineering, Harbin Institute of Technology, Harbin, China. His current research interests include machine learning and its applications to PolSAR image interpretation.
\end{IEEEbiography}
\begin{IEEEbiography}[{\includegraphics[width=1in,height=1.25in,clip,keepaspectratio]{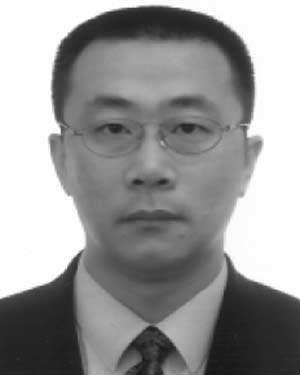}}]{Bin Zou} (M'04) received the B.S. degree in electronic engineering from Harbin Institute of Technology, Harbin, China, in 1990, the M.Sc. degree in space studies from the International Space University, Strasbourg, France, in 1998, and the Ph.D. degree in information and communication engineering from Harbin Institute of Technology, in 2001. From 1990 to 2000, he was with the Department of Space Electro-Optic Engineering, Harbin Institute of Technology. From 2003 to 2004, he was a Visiting Scholar with the Department of Geological Sciences, University of Manitoba, Winnipeg, MB, Canada. He is currently a Professor and Vice Head with the Department of Information Engineering, Harbin Institute of Technology. His research interests include SAR image processing, polarimetric SAR, and polarimetric SAR interferometry.
\end{IEEEbiography}
\begin{IEEEbiography}[{\includegraphics[width=1in,height=1.25in,clip,keepaspectratio]{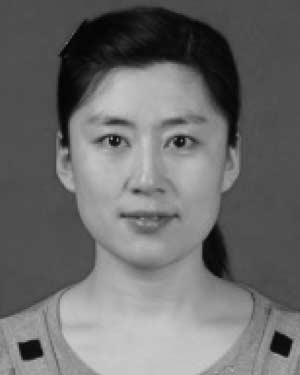}}]{Lamei Zhang} (M'07) received the B.S., M.Sc.,
and Ph.D. degrees in information and communication engineering from Harbin Institute of Technology, Harbin, China, in 2004, 2006, and 2010, respectively. Currently, she is an associate professor with the Department of Information Engineering, Harbin Institute of Technology. She serves as the Secretary of IEEE Harbin Education Section. Her research interests include remote sensing images processing, information extraction and interpretation of high-resolution synthetic aperture radar, polarimetric SAR, and polarimetric SAR interferometry.
\end{IEEEbiography}

%
%




\end{document}